\documentclass{article}
\usepackage{arxiv}

\usepackage{amsmath}
\usepackage{amsthm}
\usepackage{amssymb}
\usepackage{graphicx}
\title{Human-interpretable clustering of short-text using large language models}

\author{
 Justin K. Miller \\
  School of Physics\\
  University of Sydney\\
  Camperdown, NSW 2006 \\
  \texttt{justin.k.miller@sydney.edu.au} \\
   \And
 Tristram J. Alexander \\
  School of Physics\\
  University of Sydney\\
  Camperdown, NSW 2006 \\
  \texttt{tristram.alexander@sydney.edu.au} \\
}

\begin{document}
\maketitle

\begin{abstract} 
Clustering short text is a difficult problem, due to the low word co-occurrence between short text documents. This work shows that large language models (LLMs) can overcome the limitations of traditional clustering approaches by generating         \newline   embeddings that capture the semantic nuances of short text.  In this study clusters are found in the embedding space using Gaussian Mixture Modelling (GMM).  The resulting clusters are found to be more distinctive and more human-interpretable than clusters produced using the popular methods of doc2vec and Latent Dirichlet Allocation (LDA).  The success of the clustering approach is quantified using human reviewers and through the use of a generative LLM.  The generative LLM shows good agreement with the human reviewers, and is suggested as a means to bridge the `validation gap' which often exists between cluster production and cluster interpretation.  The comparison between LLM-coding and human-coding reveals intrinsic biases in each, challenging the conventional reliance on human coding as the definitive standard for cluster validation.

\end{abstract}

\section{Introduction}
Short text is playing an increasingly important role in human expression and interaction, due to the widespread use of social media platforms involving short text such as X (formerly Twitter), Weibo, WhatsApp, Instagram and Reddit.  The enormous quantities of data produced by users of these platforms holds the promise of not just real-time identification of events~\cite{atefeh2015} and current opinions~\cite{ravi2015}, but also a deeper understanding of the drivers of information flow between users on these platforms~\cite{pei2014}.  A typical first step in engaging with the large data sets is to reduce the complexity, by clustering the text data into similar groups~\cite{gudivada2018}.  However, short text clustering is challenging, due to the limited contextual information available in a single piece of text, 

and the low incidence of word co-occurrence between short texts~\cite{hong2010,cheng2014}. Given the possible applications of reliable short text clustering, there has been significant focus on this problem in the machine learning community, with an increasing number of methods developed to provide a deeper understanding of large collections of short text data~\cite{ahmed2022short,qiang2022}. However, a major criticism of current clustering approaches is that the resulting clusters are often difficult for humans to interpret~\cite{doogan2021}.  In addition, automated metrics used to quantify clustering success appear to be poorly correlated with interpretability~\cite{chang2009}.  There is thus a two-fold challenge, the development of an approach which can lead to interpretable clusters of short text data, and an automated approach to stand as a proxy for humans interpreting the resulting clusters.
This paper aims to close the `validation gap' when creating clusters of short text.  As such it has two sub-aims: to identify an approach for generating human-interpretable clusters of short text; and to identify ways to validate clustering success in terms of human-interpretability.

We quantify interpretability by using both human reviewers and a second generative LLM to interpret the clustering results.  We show that using an LLM in the clustering process produces more interpretable clusters than leading competing methods, and that interpretation using a generative LLM is a reliable proxy for human reviewers.  We outline a general approach which may be used to automate the short-text clustering problem while also providing a measure of clustering success.  This approach addresses the `validation gap' that has plagued automated topic modelling approaches, in which automated metrics often indicate success when human reviewers do not~\cite{hoyle2021}.  Despite the range of topic modelling approaches available~\cite{churchill2022evolution}, there are only a small number that see widespread use~\cite{ahmed2022short}.  We compare an LLM-based approach, using a mini-language model (MiniLM) , with the leading topic modelling approach, latent Dirichlet allocation (LDA)~\cite{blei2003}, and an embedding method, doc2vec~\cite{le2014}, that was found to outperform other clustering methods when using a gold-standard comparison~\cite{curiskis2020}.  

We focus specifically on a short text data set consisting of `Twitter bios', which contains user responses to the general prompt `Describe yourself'.  Users have 160 characters to respond to this prompt, resulting in highly variable short text.  We begin by using the three methods of LDA, doc2vec+GMM (henceforth `doc2vec') and an LLM that specifically generates embeddings from text all-MiniLM-L6-v2~\cite{wang2020minilm, reimers-2019-sentence-bert} + GMM (`MiniLM') to find clusters in this data set. This particular LLM was used as it gives high performance across different metrics while being fast to run on an office PC, with only 384 dimensions when it vectorises text~\cite{reimers-2019-sentence-bert, MTEBLeaderboardHugging}.  We then ask human reviewers to interpret and rate the resulting clusters.  We examine what constitutes a `good' cluster in light of the cluster characteristics and the human reviews, and look for possible metrics that can provide a quantitative measure of success.  We then turn to the possibility of using a generative LLM, ChatGPT, to act as a proxy for the human reviewers.

\section{Related Work}

\subsection{Topic-modelling and document embedding methods}

Topic modelling involves looking for topics that span the range of documents in a corpus~\cite{qiang2022}.  Typically this involves building a probabilistic model based on observed word frequencies~\cite{blei2003}, and then for short text, assuming that each document belongs to only one topic~\cite{qiang2022}.  Documents in a given topic are then taken to form a cluster~\cite{curiskis2020}. Latent Dirichlet allocation is commonly used due to its ease of use~\cite{jelodar2019}, however the severe sparsity of short text data is known to lead to poor performance~\cite{cheng2014}.

Embedding approaches instead seek to represent the text data in a vector space~\cite{ahmed2022short}, where standard clustering techniques such as k-means or Gaussian mixture modelling can be applied~\cite{rodriguez2019}.  {In this class the doc2vec method~\cite{le2014}, which establishes an embedding space by training on the data set of interest, has been found to be successful at producing clusters which agree with prior assigned labels (i.e. agreeing with a `gold standard')~\cite{curiskis2020}.}

An embedding space created using the data set of interest is in contrast to embeddings created using LLMs, as typified by the approach of Bidirectional Encoder Representations in Transformers (BERT)~\cite{devlin2019,subakti2022} and other generative LLMs~\cite{WOS:001242193200001}.  In an LLM such as BERT, the embedding space is built by training on a large external data set.  The data set of interest is then embedded in this established vector space.

Another key distinction between LLM's and language models created by doc2vec or Long Short Term Memory (LSTM), is the size of the resulting model~\cite{bender2021dangers}. The transformer-based architectures underpinning LLM-based models typically have billions of parameters and are trained on hundreds of GB of text data~\cite{floridi2020}.  Doc2vec in contrast involves only the data set of interest, and has been shown to require a large dataset to achieve high quality embeddings \cite{lau-baldwin-2016-empirical}. In addition, the architecture of non-transformer based neural networks means that these types of language models show diminishing returns past a certain training set size~\cite{bender2021dangers}.

\subsection{Clustering through Gaussian Mixture Models}

Once the embedding space is constructed, Gaussian Mixture Models (GMM) can be used to cluster document embeddings by modeling the distribution of the data as a mixture of a user-specified number of Gaussian distributions, each representing a cluster. Unlike k-means, which assumes spherical clusters, GMM allows for clusters of varying shapes and sizes, making it more adaptable to the complex structures found in text embeddings~\cite{patel2020clustering}. This flexibility is particularly beneficial when working with semantic data, as document embeddings often capture rich and nuanced relationships between words and topics\cite{sia-etal-2020-tired}.  GMM treats each document's embedding as a data point in a high-dimensional space. When applying a GMM to this data each document (i.e. point in the space) can then be assigned to a cluster based on the probability that it belongs to a particular Gaussian component. Like k-means, the number of clusters must be set at the outset as a model parameter, with changes in the cluster number leading to different results~\cite{fraley1998many}.

\subsection{Metrics of interpretability and distinctiveness}

The problems facing the use of automated metrics have been widely identified~\cite{chang2009,lipton2018,doogan2021,hoyle2021,qiang2022}, with the popular metric of coherence singled out for often not aligning with how humans rate clusters~\cite{doogan2021,hoyle2021}. Therefore it would be appear to be unwise to rely only on automated metrics to determine the success of a clustering method, and human intepretability should also be considered.
A key measure of interpretability is the ability of humans to easily name a cluster~\cite{doogan2021} and see agreement between top words of a cluster and sample documents. 
A complication which arises when framing success in cluster creation using human interpretability, is that humans have biases and limitations when faced with the tasks of clustering (also called categorization, the identification of unlabelled groups of similar objects) and classification (labelling of already existing clusters)~\cite{anderson1991}.  Humans appear to cluster preferentially along only one selected dimension in high dimensional data~\cite{doan2023}, and the dimension chosen depends on the experiences and biases of an individual performing the clustering~\cite{broker2022}.  Human-labelled data is typically taken as a base truth, or gold standard, however these limitations suggest that, at least for cluster labels, there may be no such thing as a gold standard.  One of the goals of this work is to determine a measure of clustering success in the presence of this ambiguity.

While interpretability has been identified as important, we suggest that clusters should also be distinctive.  A subset of a randomly generated cluster may appear coherent, and even interpretable, to a human reviewer, due to the well-known human bias of seeing patterns in noise (apophenia)~\cite{clarke2008}.  However, when dealing with large randomly-generated clusters, any fluctuations are expected to average out and the clusters will appear similar to each other.    
We need a measure to capture this degree of distinctiveness between clusters. To do so, we make use of the method of keyword analysis from corpus linguistics. Keyword analysis can identify words that occurred more frequently in specific clusters compared to a reference corpus, as outlined by Baker, creating a quantifiable way to assess how linguistically different the clusters are from the corpus (2006)~\cite{baker2006using}.

\section{Methods}

\subsection{Data sets and data cleaning}

The data used in this paper consists of 38,639 Twitter user bios from users who used the words `trump' or `realDonaldTrump' between 3rd September and 4th September 2020. This limits the domain of users to those with some interest in US politics. The user timelines of these users were collected from Twitter using the Twitter API v2 between September 6th 2020 and September 16th 2020. From these user timelines the username and bio for every user was extracted, however only the bio was used in the clustering.

Using the dataset, clustering models were developed using Python 3 to explore patterns within the bios. For data preprocessing, emojis were converted to their CLDR Short Name using the Emoji package~\cite{kimCarpedm20Emoji2023}. For both the LDA and Doc2vec model, stopwords were removed and words were lemmatized with NLTK 3.6.5~\cite{loper2002nltk}.
% changed here to be clearer
As a key component of Large Language Models is the attention component, removal of stop words and lemmatization is likely to negatively affect model performance and so was not done.
LDA and doc2vec models were built using Gensim 4.1.2~\cite{rehurek2011gensim}. The Doc2vec method used the parameter values suggested by Curiskis et al. ~\cite{curiskis2020}: 100 dimensions and 75 epochs.    

A Gaussian mixture model with diagonal covariance was then applied to the embeddings generated by the doc2vec and MiniLM models. Diagonal covariance was chosen as each dimension was mutually orthogonal. The chosen number of clusters and topics (K) was set to 10. Ten clusters was chosen as a balance between complexity (number of clusters to identify) and distinctiveness (how well-defined each cluster is).
 
A null model was also created by randomly assigning every bio to a cluster.  This random model serves as a baseline to compare with the other clustering methods.  The data and code used in this paper can be found at https://github.com/justinkmiller/BioData.

\subsection{Validation through human review}

To evaluate the clustering models, human reviewers (N=39) were recruited through Prolific\cite{Prolific} and directed to Redcap at the University of Sydney\cite{harris2009redcap, harris2019redcap}.
Reviewers, who were American and native English speakers, evaluated clusters generated by LDA, doc2vec, MiniLM, and the randomly created clusters. For the LDA model, samples included the top 10 high-probability words and 20 bios assigned to each topic. From the doc2vec, MiniLM, and random models reviewers were provided with the top 10 most frequently used words in each cluster and 20 randomly chosen bios from each cluster.  Reviewers were shown each cluster from one of the 4 models in a random order, and were asked the following questions: 

\begin{itemize}
\item ``Create a name using less than 10 words to summarize the top 10 words/emojis and the sample bios of the Sample Cluster. If you believe it is not possible to do this with a cluster, write `None'."
\item ``When you named a Sample Cluster, were you confident that the name summarized the whole cluster?" Reviewers can answer this question by selecting one of the following: Not at all Confident/Not Confident/Neutral/Confident/Very Confident. 
\vspace{5mm}

For the following questions, reviewers can respond with: Strongly Disagree / Disagree / Neutral / Agree / Strongly Agree

\item ```The situation when the top words/emojis of a cluster fit together in a natural or reasonable way' Do you agree that the above statement describes the top words/emojis of this Sample Cluster?" 
\item ```The situation when the Sample bios of a cluster fit together in a natural or reasonable way' Do you agree that the above statement describes the sample bios of this Sample Cluster?" 
\item ```The Top Words/Emojis provide an accurate summary of the sample bios' Do you agree that the above statement describes this Sample Cluster?" 

\end{itemize}
% Are these figure numbers still correct?
See Fig~ S10 and S11 to see how this was presented to the reviewers.

Survey responses were scaled from 1 (``Strongly Disagree" or ``Not at All Confident") to 5 (``Strongly Agree" or ``Very Confident"). As a Likert scale was used, ordinal regression was used to compare the methods~\cite{burkner2019}.
The study employed four Bayesian cumulative ordinal regression models, with the dependent variable being the questions the reviewers were asked and the independent variable being the cluster-creating model. All ordinal regression models were created using the BRMS package in R~\cite{brms2021}. Four chains were used, with 5,000 iterations, 2,500 of which were warm-up to give the model sufficient time to converge to a posterior distribution. This resulted in a total of 10,000 draws from the posterior distribution for each of the four models. The priors for each model were the default from the BRMS package: student's t-distribution with 3 degrees of freedom, location parameter 0, and scale parameter 2.5. These models generate a latent variable \^{Y}~\cite{burkner2019}, with thresholds for each response level. A higher \^{Y} indicates a greater likelihood of ``Agree" or ``Strongly Agree". Model performance is evaluated by comparing the coefficient distributions against the random model, which is used to define the location of 0 on the $x$-axis.

\subsection{Automated clustering metrics}

Metrics such as coherence and distance were calculated to assess the quality of the clusters.
Coherence was measured using the $C_{UMASS}$ and $C_V$~\cite{rehurek2011gensim} methods after removing all stopwords and punctuation with NLTK 3.6.5~\cite{loper2002nltk}. The top 10 words for each cluster were used in the calculation of coherence.

For all four models, the embeddings were generated by all-MiniLM-L6-v2~\cite{wang2020minilm, reimers-2019-sentence-bert} to compute distance- and position-based metrics. The silhouette score is a commonly used metric in cluster evaluation that measures how similar an object is to its own cluster compared to other clusters \cite{9260048}. It is defined as:
\begin{equation}
\label{eq:silhouette}
{s(i) = \frac{b(i) - a(i)}{\max(a(i), b(i))}}    
\end{equation}
where \(a(i)\) is the average intra-cluster distance, and \(b(i)\) is the average nearest-cluster distance for point \(i\).

The distance used in the calculation of silhouette scores within this paper was the cosine distance, as cosine distance is often used within text embedding research \cite{pan2020towards, toshevska2020comparative, bamler2017dynamic}, and euclidean distance and cosine distance have been found to give similar results in high dimensional spaces \cite{10.1145/967900.968151}.

 Subsequently, we computed the mean silhouette score for the bios in each respective cluster (using sklearn~\cite{scikit-learn}), providing an aggregated measure of their cohesion. Cluster distance was calculated by averaging over all elements of a cluster to find the cluster centroid in the embedding space, and then computing the pairwise distance between the cluster centroids.  The shortest distance for each cluster centroid was then taken as a measure of how well the bios in that group were clustered. As GMM gives a standard deviation for each cluster and each dimension, comparing the average standard deviation across dimensions for each cluster can give a measure of cluster width. It can be assumed that in an embedding space, a smaller width of a cluster means that it is more precise and thus easier for a human to interpret.  In this work, this metric is referred to as mean standard deviation.

To examine the correlation between reviewer responses and automated metrics, Spearman rank correlation was calculated for each question across all 40 clusters and $N_R = 39$ reviewers. The Spearman rank correlation, a non-parametric measure, assesses the strength and direction of association between two ranked variables. This gives a distribution of 39 reviewers and their correlation with the automated metrics. Spearman rank correlation was used since the data is ordinal and a rank-based approach gives the most accurate representation of the data \cite{ornstein2016asymptotic}.

A list of keywords associated with each cluster was also created using an automated method: the frequency of words within clusters was compared to their frequency in the entire corpus, and the Bayes factor was calculated for each word, with words exceeding a Bayes factor of 10 classified as keywords. This threshold was used to indicate if a word is significantly more prevalent in a particular cluster compared to its general frequency in the entire corpus~\cite{wilson2013embracing}.

\subsection{Cluster name metrics}

For clarity's sake, each cluster was given a name by the authors to identify it among all 40 clusters. A name was created by looking at the names human reviewers gave a cluster, the top 10 words of each cluster, and reviewing 40 - 50 bios from each cluster.  If no meaningful name could be created then the cluster would be designated ``U" for ``Undecided" followed by a number.  The number was added as an identifier, to provide a unique label for each cluster. Clusters within the same model that seemed to warrant the same name (e.g. `Left leaning') were given a number (e.g. 1 or 2) to make the label unique.

The consistency, $S$, of naming by reviewers is used as a quantitative measure of `interpretability'.  
Using the linguistic terminology of tokens (word instances), $w$, and types (the different types of words used) $W$, 
the words used by reviewers to describe cluster $i$ form a corpus of $N_w$ tokens $w_i$ and a corpus of $N_W$ types $W_i$. 
Consistency of naming is then defined by counting the number of instances of a given type $W_{i,j}$ in token corpus $w_i$, 
and then normalizing by the number of names created for each cluster $N_R$ (e.g. the number of human reviewers). 
Let $w_{i,k}$ be the $k$th token in corpus $w_i$ and $W_{i,j}$ the $j$th type in corpus $W_i$, then the consistency $S_{i,j}$ of type $W_{i,j}$ is defined as:
\begin{equation}
{S_{i,j} = \frac{\sum_{k=1}^{N_w} \delta(w_{i,k},W_{i,j})}{N_R}}
\label{eq:consistency}
\end{equation}
{where $\delta(w_{i,k},W_{i,j})$ is the Kronecker delta function defined as:}
\begin{equation*}
{\delta(w_{i,k},W_{i,j}) = 
\begin{cases} 
1 & \text{if } w_{i,k} = W_{i,j} \\ 
0 & \text{if } w_{i,k} \neq W_{i,j}. 
\end{cases}
}
\end{equation*}

%\begin{equation}
%{S_{i,j} = \frac{\sum_k (w_{i,k} = W_{i,j})}{N_R}}
%\label{eq:consistency}
%\end{equation}

The words corresponding to the top five values in $S_i$ are displayed to indicate the most prominent words used to describe the $i$th cluster.  
Interpretability $I_i$ of cluster $i$ is taken to be the consistency of the most frequently used type (i.e. the top word) in the cluster names:

\begin{equation}
{I_i = \max_{j=1}^{N_W}(S_{i,j})}
\label{eq:interpretability}
\end{equation}

As discussed earlier, interpretability is not enough to define successful clustering: clusters should also be distinctive.  
A useful measure for distinctiveness is the Jensen-Shannon Divergence (JSD)~\cite{lu2021diverging}.  
JSD is used to measure the similarity between two corpora based on the probability distributions of their word frequencies:
\begin{equation}
{JSD_{1,2} = \frac{1}{2} D_{KL}(P_{C_1} \parallel M) + \frac{1}{2} D_{KL}(P_{C_2} \parallel M)}
\label{eq:JSD}
\end{equation}
where \(P_{C_1}\) and \(P_{C_2}\) are the probability distributions of word frequencies in corpora \(C_1\) and \(C_2\), respectively, 
and \(M = \frac{1}{2}(P_{C_1} + P_{C_2})\).  
The Kullback--Leibler divergence \cite{kullbackInformationSufficiency1951} $D_{KL}(P_{C_j} \parallel M)$ between two probability distributions $P_{C_j}$ and $M$ is defined as:

\begin{equation}
D_{KL}(P_{C_j} \parallel M) = \sum_{i} P_{C_j}(i) \log \frac{P_{C_j}(i)}{M(i)}
\end{equation}

In this work distinctiveness is measured by computing the JSD between corpora after stemming all words to their base form.  
A `distinctiveness' measure $D_i$ for cluster $i$ is introduced based on the JSD:
\begin{equation}
{D_i = \min_{j \ne i}^{N_c}JSD_{i,j}}
\label{eq:distinctiveness}
\end{equation}
where the JSD for a cluster $i$ is given by the smallest JSD between this cluster and all other clusters~\cite{menendez1997jensen} (where $N_c$ is the total number of clusters).  
For a given model, the average and standard deviation of the $D_i$ are computed.  
A higher average JSD suggests that the clusters in the model cover a broader range of distinct topics and exhibit more linguistic variety.  
%Additionally, to test how reliably someone can name a cluster the same way, thus whether it can be regularly interpreted, 
%we find the fraction of cluster names that contain the most frequent word.

\subsection{Large language model name creation}

While LLMs have shown early promise for text clustering~\cite{subakti2022}, they have seen spectacular recent success in text generation, thanks to models such as ChatGPT~\cite{openai2023}.  ChatGPT has been rapidly embraced by the public, but generative LLMs more broadly have also seen applications in areas as diverse as 
finance~\cite{huang2023}, health~\cite{si2019}, law~\cite{chalkidis2019} and academia~\cite{altmae2023}.  While the focus of these models has been on producing human-like content~\cite{liu2023}, more quantitative applications are starting to emerge, such as sentiment analysis~\cite{susnjak2023} and text annotation~\cite{gilardi2023}. More recently, LLMs have been shown to replicate human reasoning when analysing text~\cite{aher2023using} which raises the prospect that they might also be used as a proxy for a human seeking to interpret resultant clusters, and provide an avenue for a metric which correlates with human interpretability.

To explore the possibility of automating the human review process, ChatGPT was used to perform all the tasks asked of the human reviewers.  Using ChatGPT's API \cite{openai_2023_gpt4}, ChatGPT was instructed to create a name using up to 5 words for a given cluster.  As with the human reviewers, ChatGPT was provided with the top 10 words of each cluster and a random sample of 20 Twitter bios.  In cases where ChatGPT deemed a cluster unnameable or just a random amalgamation, it was instructed to return `None'. This procedure was repeated 39 times to match the number of human reviewers.  The same measures applied to the human reviewers were then applied, to provide measures of interpretability and distinctiveness of the ChatGPT-produced cluster names.

\section{{Results}}
\subsection{{Human Cluster Evaluation}}
\begin{figure}[htbp]
 \begin{flushleft}
  \includegraphics[width=0.9\linewidth]{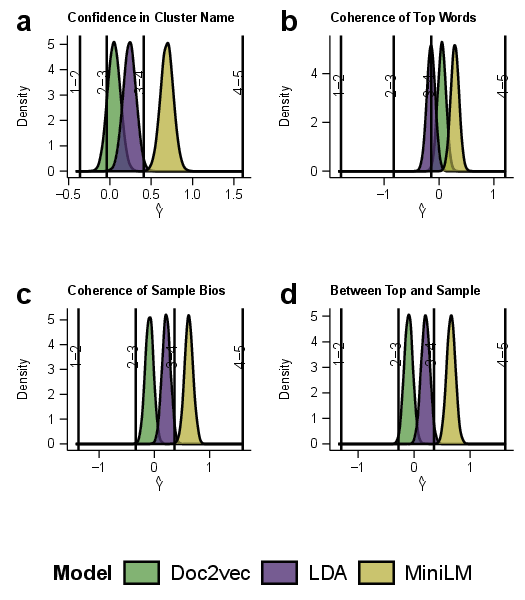}
  \caption{Ordinal Regression Analysis of Clusters created by LDA, Doc2vec, and LLM illustrating the outcomes of an ordinal regression model applied to the ratings of 39 reviewers. The reviewers assessed Twitter bio clusters and were asked to rate the coherence of the clusters across four categories: (a) Confidence in Cluster Name, (b) Coherence of Top Words, (c) Coherence of Sample Bios, and (d) Coherence between the top words and sample bios. Each panel (a-d) represents the probability density function ($\hat{Y}$ ) of ratings in each category, showcasing the statistical modeling of ordered categorical data. The reviewer ratings of the random clusters are set to 0 so any value greater than 0 is performing better than random. The vertical lines identify the transitions between values on the likert scales used by the reviewers.  We see that MiniLM has consistently been scored at 4 on all measures.}
  \label{fig:Ordinal Regression}
  \end{flushleft}
\end{figure}
We begin by quantifying the cluster evaluation performed by the reviewers.  Performing an ordinal regression on the referee responses, we see in Fig.~\ref{fig:Ordinal Regression} that the MiniLM has performed consistently better than the other two methods on all metrics.  LDA performed consistently better than random, while doc2vec performed consistently worse.  Inspection of the doc2vec clusters reveals that the clustering approach has indeed made use of features that are largely invisible to human reviewers, such as a cluster of bios using low-frequency words.  Interestingly doc2vec was found to be successful at clustering when a gold standard is present~\cite{curiskis2020}, but in this case there is no gold standard.  The results are consistent with the finding that traditional clustering approaches produce clusters that are often uninterpretable by humans~\cite{doogan2021}.
\begin{figure*}[tbp]
 \begin{flushleft}
  \includegraphics[width=0.9\linewidth]{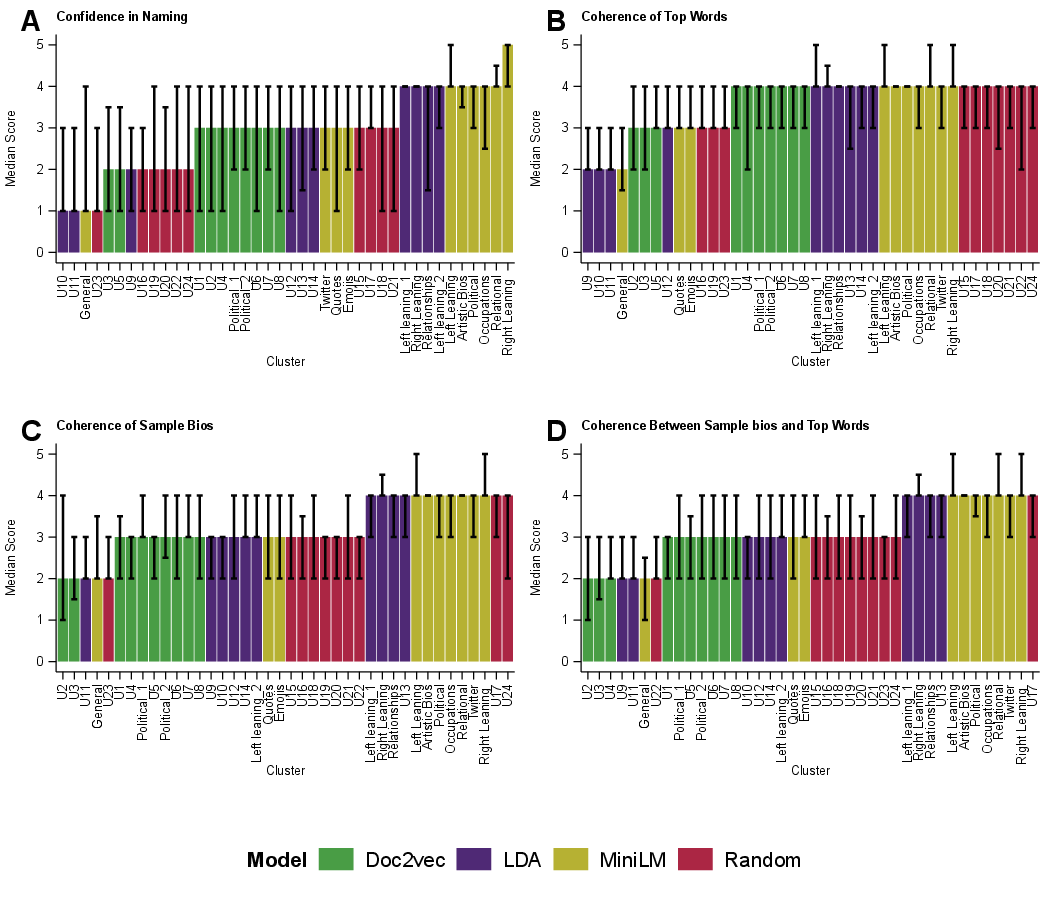}
  \caption{The median reviewer score for each cluster across the four categories: (a) Confidence in Cluster Name, (b) Coherence of Top Words, (c) Coherence of Sample Bios, and (d) Coherence between the top words and sample bios. Error bars represent the first and third quartile scores. All clusters with the U designation were unable to be named by the authors of this paper.}
  \label{fig:Median}
  \end{flushleft}
\end{figure*}

When we look at the performance by cluster in Fig.~\ref{fig:Median}, we find the results are more nuanced.  In Fig.~\ref{fig:Median}(a) we see that while the reviewers were usually confident in naming clusters produced by  MiniLM, some of the clusters were less clear, and one cluster could not be named at all.  The cluster names provided on the $x$-axis have been created by the authors based on observation of the reviewer-provided names and a deeper review of the clusters.  The `General' cluster could not be named by reviewers, and the `Twitter', `Quotes' and `Emojis' clusters were also challenging for the reviewers.  We examine the names provided by reviewers further below.

Interestingly, as seen in Fig.~\ref{fig:Median}(b), coherence of top words, as rated by the reviewers, is not a good predictor of confidence in naming, with little distinction detected between the models, and even the randomly generated clusters rated highly.  Looking at Fig.~\ref{fig:Median}(c,d), MiniLM appears to have aggregated bios such that they appear more coherent to the reviewers, as well as producing clusters with greater coherence between the bios and the top words.  This suggests that top words can play a role in naming, as is common practice with topic modelling~\cite{doogan2021}, however the high ratings also assigned to random clusters suggests that human estimations of coherence do not provide a measure of clustering success.
\subsection{{Automated Evaluations}}
\begin{figure}[htbp]
 \begin{flushleft}
  \includegraphics[width=0.9\linewidth]{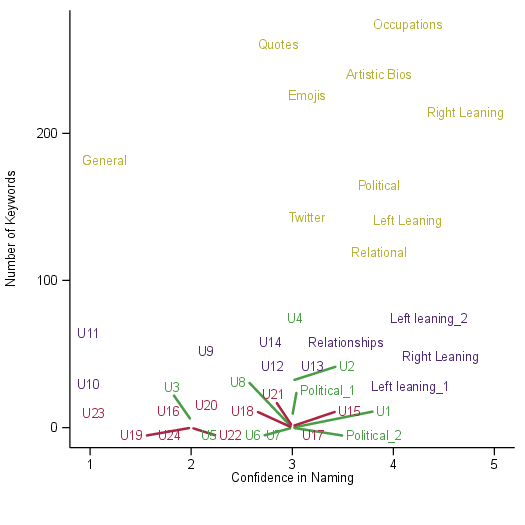}
  \caption{The number of keywords in each cluster, where keywords are determined by comparing the expected frequency of words in a cluster against their actual frequency, using a Bayesian factor greater than 10 to assess if the difference is statistically significant. The names of each cluster are the names given by the authors and the prefix U indicates that a cluster was not able to be named.  Colors identify the clustering model used and are consistent with the color code used in Fig.~\ref{fig:Median}.  The MiniLM clusters have significantly more keywords than the clusters found using the other methods.}
 \label{fig:Keyword}
 \end{flushleft}
\end{figure}

 We identify the number of keywords special to a cluster and plot this versus human confidence in naming, see Fig.~\ref{fig:Keyword}.  We find that all the clusters produced by MiniLM have large numbers of distinctive keywords, even the clusters which human reviewers were unable to interpret (such as the `General' cluster).  In contrast, all the randomly generated clusters have zero, or close to zero, distinctive keywords, as expected for a complete mixing of the data set.  The methods of LDA and doc2vec lead to some clusters with distinctive keywords, but much less distinctiveness than MiniLM approach.  We find therefore that the large language model has allowed for the creation of not just more human interpretable clusters, but also clusters which are more distinctive relative to each other.  

\begin{figure}[htbp]
  \centering
  \includegraphics[width=0.9\linewidth]{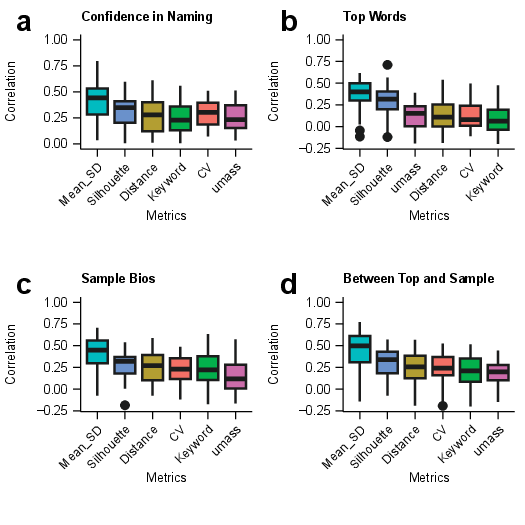}
  \caption{Boxplots showing the Spearman Rank Correlation between the reviewer provided ratings and the six automated metrics of mean cluster standard deviation, silhouette score, Euclidean distance, number of keywords, CV coherence and UMass coherence (x-axis labels), with reviewer ratings on (a) Confidence in naming; (b) Coherence of top words; (c) Coherence of sample bios and (d) Coherence between top words and sample bios. Coherence and keywords correlate poorly with reviewer ratings. Mean standard deviation appears to provide the best correlation with the ratings. The large variability in correlation across reviewers is evident, with outliers identified with solid circles, i.e., some reviewers correlated well with some measures, while others showed no correlation or negative correlation for some measures.}
  \label{fig:Metric_Human}
\end{figure}

In Fig.~\ref{fig:Metric_Human} we examine the correlation between these metrics and the ratings provided by the reviewers.  We see that the correlation is generally poor.  The Mean Standard Deviation metric, which serves as a proxy for cluster width in the embedded space, shows the best correlation with the human ratings, followed by the silhouette score.  The coherence measures show little correlation with the human measures, including the human ratings of coherence.  We find that these coherence measures also show poor correlation with each other (see Fig.~S1), so failing a posited criterion of interpretability, that different coherence measures should correlate~\cite{doogan2021}.  Our results are therefore consistent with findings elsewhere that, despite being widely used, the metric of coherence is not useful when measuring human interpretability of short text~\cite{hoyle2021}.  We note that the number of keywords also does not correlate with the human measures, however as discussed earlier, we argue this metric provides a measure orthogonal to interpretability, and a lack of correlation is consistent with this.

\subsection{{Cluster Names Evaluation}}
\begin{figure*}[htbp]
 \begin{flushleft}
  \includegraphics[width=0.9\linewidth]{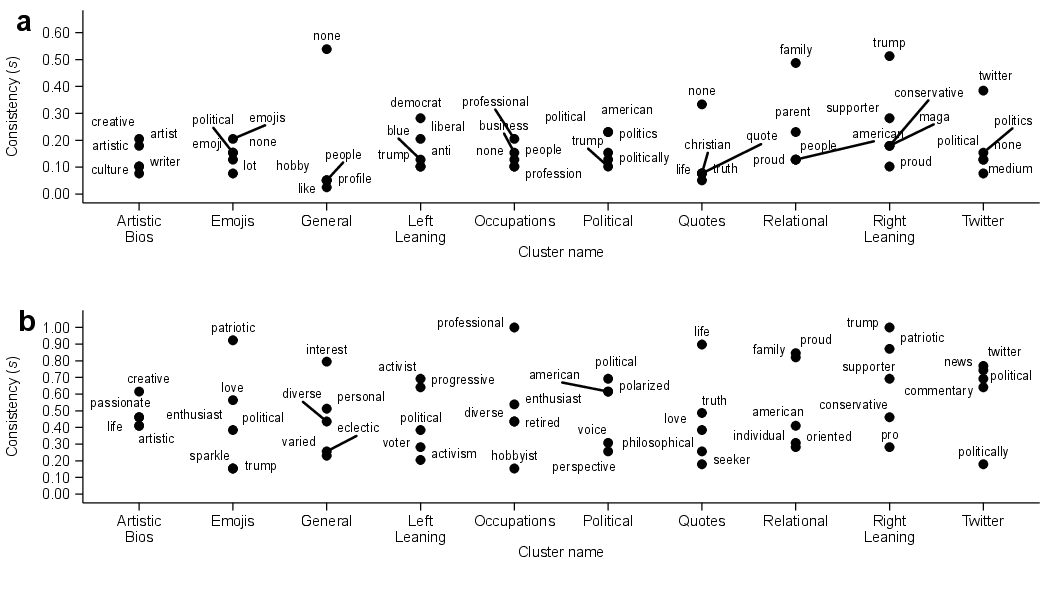}
  \caption{The five words with the highest Consistency ($S$) used by (a) reviewers and (b) ChatGPT to name the clusters created by MiniLM. Along the x-axis are the names given to each cluster by the authors of this paper.  We see that the words used to describe the clusters are largely consistent between ChatGPT and the reviewers, however there are cluster-dependent distinctions revealing human and machine limitations as discussed in the text.}
  \label{fig:Reviewer_Name}
  \end{flushleft}
\end{figure*}

In Fig.~\ref{fig:Reviewer_Name}(a) we show the most popular words appearing by cluster in the reviewer LLM cluster names.  The x-axis tick labels are again the researcher-given names for the associated clusters. We see that the Right Leaning cluster has been consistently named by reviewers to be a `Trump' related cluster, and the Relational cluster is similarly annotated by the word `family'. 
  
One of the goals of our work is to validate a proxy for human interpretation.  To explore the possibility of a generative large language model replicating human interpretation of a cluster, we contrast the reviewer names, Fig.~\ref{fig:Reviewer_Name}(a), with those produced by ChatGPT-4.0, Fig.~\ref{fig:Reviewer_Name}(b).  We find that ChatGPT provides names for all the clusters, unlike many of the human reviewers who selected `None' for some of the clusters.  The names ChatGPT has given look consistent with the names provided by the reviewers, and those used by the researchers.  `General' and `Quotes' have both been given appropriate names.  This suggests that the human reviewers struggled to identify an underlying similarity in these clusters of bios.  This is consistent with a bias identified in human classification, in which humans tend to seek a low dimensional representation for data when they are asked to perform a classification task (e.g. choosing to focus only on expressions of political ideology)~\cite{broker2022}.  `Quotes' do not easily fit into this representation, although we note that some reviewers did detect this feature in the set of bios.  The `General' class has been picked up by ChatGPT, and such a class is known to exist in expressions of human identity~\cite{mackinnon2010}, but this class was overwhelmingly left unnamed by the reviewers.  Interestingly, the `Emojis' cluster was named by ChatGPT according to its content rather than the medium.  This is perhaps unsurprising, given the tool focuses on word usage, and so has provided names based on the content rather than the form of the content.  We conclude therefore that clusters that weren't named by the reviewers, could be named by ChatGPT, suggesting human bias rather than limited information may be involved, however we also found that when faced with finding a `meta' name for a cluster, the humans performed better than ChatGPT.

\begin{figure}[htbp]
 \begin{flushleft}
  \includegraphics[width=0.9\linewidth]{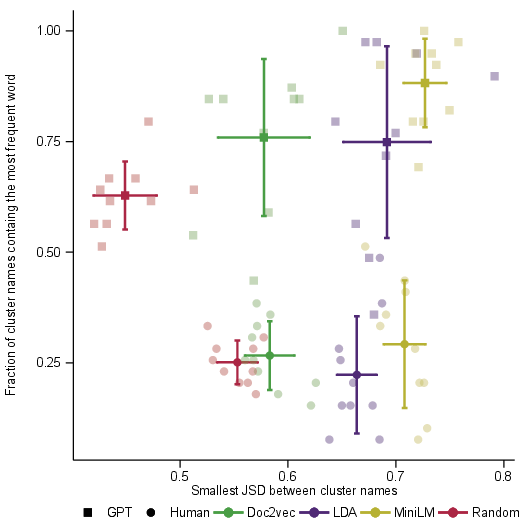}
  \caption{The interpretability (given by Eq.~(\ref{eq:interpretability})) and distinctiveness (given by Eq.~(\ref{eq:distinctiveness})) for each cluster as determined by the names given by human reviewers (circles) and ChatGPT (squares) for the four different clusterings (Doc2Vec, LDA, Random and LLM).  The fainter symbols are results by cluster, the more solid symbols with error bars indicate the averages.  {MiniLM produces clusters that are able to be named more consistently and have a greater separation between them than the clusters produced by the other methods}. ChatGPT is broadly consistent with the human reviewers, though with significant differences, as discussed in the text.}
  \label{fig:Interp_diversity}
  \end{flushleft}
\end{figure}

We now take a quantitative approach to comparing the names given by ChatGPT and our human reviewers, comparing the interpretability and distinctiveness of their names.
We see in Fig.~\ref{fig:Interp_diversity} that humans and ChatGPT are broadly consistent with each other. 
The MiniLM clusters are given the most distinctive names and are also the most consistently interpreted.  We see however some interesting differences between the human and machine approaches.  ChatGPT is more consistent in its naming between runs, which is unsurprising given it is a single large language model as compared to a set of different human reviewers.  This leads to a set of higher interpretability scores when compared to the human scores.  However, while MiniLM has scored well in both absolute and relative terms using ChatGPT, the random clusters have also performed well in absolute terms.  This is likely a consequence of the underlying political skew to the dataset.  ChatGPT appears to have better identified this underlying bias in the dataset, by producing lower scores in the distinctiveness measure for the random clusters when compared with the scores given by the human reviewers.  An interesting future direction would be to consider datasets with less of an underlying signal.

\section{Discussion}
  
Our work reveals an interesting contrast between reviewer confidence in naming, and the interpretability measure based on naming consistency.  When we ask human reviewers to rate their confidence that the name they have provided for a cluster is representative of the cluster as a whole, we see that MiniLM-based clusters performed significantly better than the other clustering approaches.  However, when we look at consistency between names as provided by the reviewers, we see that the methods are rated more closely, though MiniLM clusters again performed the best.  This observation is consistent with the findings by Doogan and Buntine, that looking at the results of reviewers completing a labelling task is more informative than asking reviewers about their ability with the task~\cite{doogan2021}.  Focusing on labels also allows us to develop an automated approach based on ChatGPT.  Asking ChatGPT to rate its confidence in naming would seem to be of questionable value, but asking ChatGPT to repeatedly provide a name allows us to obtain a measure of interpretability, using the tool's strength in text comprehension.  The top-word consistency measure we have introduced can be seen as a proxy for the inter-coder reliability measure suggested as a means to quantify interpretability~\cite{doogan2021}.

A limitation of using top-word consistency to measure interpretability is that it does not allow for words that are synonymous to be aggregated, and thus could underestimate how consistent reviewers are in interpreting a cluster. A figure such as \ref{fig:Reviewer_Name} helps to check for this, by revealing if there are competing similar words.  For instance we see examples of words that are distinct due to their choice of form (`political', `politics' and `politically' for the Political cluster), which if stemmed and so combined would lead to a higher consistency score. 
 However, we also see similar words that would not be combined by stemming, for instance `creative' and `artistic' describing the `Artistic Bios' cluster or `democrat' and `liberal' describing the `Left Leaning' cluster.  An approach to manage this variety could be to embed the words using a large language model and examine the width of the resulting cluster (e.g. by computing the mean standard deviation).  We have chosen to use a simple and transparent approach here, but a method to account for semantic overlap between different words would be an interesting direction to explore.

We have chosen to base our distinctiveness measure on names provided, however an alternative is to use the complete cluster information, as was done in our keyword analysis.  Complete information better distinguishes randomly created clusters and the separation of clusters in a semantic space, however a measure based on names better reflects the human perception of the clusters.  We have chosen this human-centric approach here, however all the documents in a cluster could be combined and a text dissimilarity measure used to quantify the distinctiveness of the clusters~\cite{shade2023}.

We see that ChatGPT produces results that are consistent with human reviewers, though with some interesting distinctions.  ChatGPT found greater variability between the clustering approaches, and appeared to be significantly better at distinguishing the random clusters than the human reviewers.  This suggests that ChatGPT is less prone to the biases known to affect humans when naming clusters, such as the temptation to see patterns in noise.  However, we should note that the data set we used had a strong political orientation, such that reviewers naming random clusters as `political' is reflective of an underlying signal rather than noise.  This signal however is present in all clusters, and ChatGPT appears to have been more consistent in detecting this inherent bias and so overall giving a lower distinctiveness measure to the random clusters than that given by the human reviewers.  These results indicate the importance of including a null model when building and interpreting clusters.  Observed in isolation, all ratings appear to suggest that the clusters are cohesive and interpretable.  It is only when comparing models, and more particularly when comparing with a random model, that the success in clustering can be identified.

The human difficulty of distinguishing a signal from noise is the reason we haven't implemented a commonly suggested approach for measuring interpretability: topic or word intrusion~\cite{chang2009}.  The combination of short text and the inherent ambiguity in the clusters due to the absence of a gold standard, make it difficult for a human reviewer to identify an `intruder' whether at the topic or document level.  This is particularly the case with short text bios.  The information provided by a user may conceivably belong to multiple clusters, with the optimal cluster difficult for a human to determine. For instance, a bio might have a clear political orientation and clear references to family.  This ambiguity reflects the multifaceted nature of human identity~\cite{mackinnon2010}.  However to be of value, the clusters that are found should reflect some deeper divide in the data.  We should note that variability in human reviews may also depend on the choice of reviewer pool.  For this work we chose to use non-expert reviewers, however for a specialist task there may be significantly less variability in the naming.  It would be interesting to see if domain-experts perform more similarly to the large language model.

%Perhaps something about non-expert reviewers, and where an LLM might sit relative to experts

%Removed Identity component

We see that the large language model approach to text clustering has revealed interesting, and human-interpretable, clusters.  Based on the immense training sets underpinning such models, more sophisticated patterns have been found, such as the use of quotes or emojis, in addition to more clearly identifiable classes of human self-identity.  These clusters also appear to show expected relationships with each other when projected into a two-dimensional space (see Fig.~S2), with occupations sitting near artistic bios, but far from political bios.  The commonly used approach of LDA appears to have also performed well in producing distinctive and interpretable clusters, however it suffers from a repeatability failure.  Clusterings using different random seeds produce quite different partitions of the data, limiting the ability to make general comments about a particular observation (see Fig.~S4).  This appears to be less of a problem when using the doc2vec method (see Fig.~S5), however doc2vec produces clusters that are difficult for a human to interpret (such as grouping bios that use rare words together).    

An interesting direction for further work is to consider the effect of different embeddings on the clustering and interpretation process.  With the wide variety of large language models now available, a possible method for comparing these models quantitatively would be to use them for both clustering and interpretation of short text.  Embeddings created using commercially-available LLMs, such as ChatGPT, Gemini or Llama, would involve many more dimensions than those of the mini-LLM used for the embeddings in this work. Benchmarking different LLMs has found that more complex models produce embeddings that lead to clusters that better align with a gold standard metric \cite{keraghel2024beyond}.  Continuing in this line of experimentation, 
this may lead to greater distinctiveness in the clusters, but it may also lead to artifacts due to the high dimensional nature of the data.  The tools could be directly compared by placing the model results on a plot similar to that created in Fig.~\ref{fig:Interp_diversity}.  This would provide an alternative method for comparison to the human rating system currently used, or the method of comparison against a gold standard.  
Finally, in this work GMM has been used for performing the clustering in the embedding space, however there are other options available, such as density or hierarchical clustering \cite{10.1371/journal.pone.0210236}.  These different clustering methods could be directly compared using the distinctiveness and interpretability metrics, to determine if a particular clustering method combined with an LLM produces higher quality clusters.  Combining such an investigation with embeddings produced by different LLMs would also provide a more comprehensive study of the interplay between clustering approach and the chosen embedding space.

\section{Conclusion}
Based on our results, the recommended approach in short-text clustering is to use a large language model to obtain an embedding which can then be clustered.  For automated analysis of the resulting cluster we have found that ChatGPT has performed as well, or better, than human reviewers, as judged using a top-word consistency measure of interpretability.  Where ChatGPT performed poorly was in the identification of means of expression rather than content (i.e. meta elements such as the use of quotes or emojis), however on balance the model still performs better than human reviewers when faced with these clusters.  We suggest therefore that LLMs may be used in both the creation and validation of clusters, providing a means to close the well-identified `validation' gap that is common in cluster analysis~\cite{hoyle2021}.

Our results show the importance of including a null model in any clustering process, to allow for the distinction between a signal and noise.  We propose that in addition to the method outlined above, the process is repeated with a set of randomly created clusters.  This will provide a measure of success relative to the random case.

It is important to acknowledge a fundamental issue with using large language models: they are largely black boxes, and in the case of ChatGPT, maintained by a private company.  A risk is therefore that implementations and training sets can change over time, affecting the results.  We see this when we perform the cluster interpretation using different ChatGPT implementations (see Fig.~S7).  However, variability is also an issue with human reviewers (see e.g. the outliers in Fig.~\ref{fig:Metric_Human}), and there are additional checks that can be used to validate MiniLM clustering measures.  A quantification of cluster distinctiveness can be performed using all the clustering data, e.g. via keyword analysis, so providing a valuable metric without needing the automated interpretation step of a second large language model.  Our analysis also indicates that the top words are a good guide to the nature of MiniLM-created cluster.  These can be used to perform a human-based interpretation of a cluster as a check against names created by the generative LLM.  However, to obtain metrics of interpretability and distinctiveness at scale, we have validated that the generative LLM performs well.  We hope that this opens up a new approach to analysing short text, allowing new insights to be gained from this increasingly important means of human expression.

%%%%%%%%%% Insert bibliography here %%%%%%%%%%%%%%

\bibliography{Main_File}
\end{document}

\maketitle

\section{Correlation between metrics}
\begin{figure}[htbp]
 \begin{flushleft}
  \includegraphics[width=0.8\linewidth]{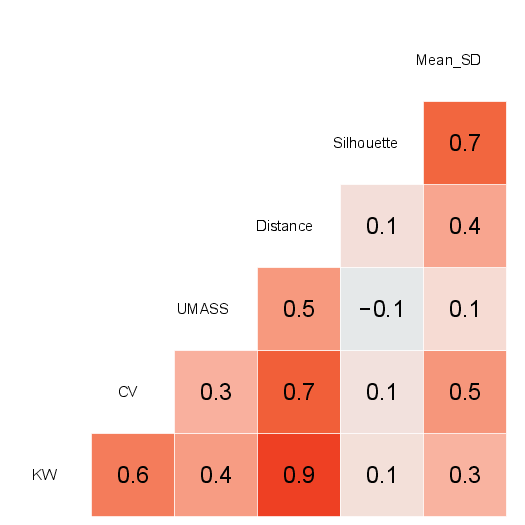}
  \caption{This figure shows the Pearson rank correlation between each of the automated metric scores for each of the 40 clusters.(KW is short for number of Keywords in a cluster)}
  \label{fig:Metric}
  \end{flushleft}
\end{figure}

To better understand how each automated metric compares, we calculated the Pearson rank correlation for each pair of metrics based on their scores within each cluster, as presented in the heatmap (see  Fig. S\ref{fig:Metric}). Notably, silhouette scores showed a correlation exclusively with the mean standard deviation. This correlation is logical given that both metrics aim to measure similar aspects. However, the silhouette scores' low correlation with other metrics is unexpected. In contrast, Keywords and Distance metrics exhibited a significant correlation, aligning with their common goal of assessing cluster distinctiveness. Despite their different approaches, this correlation validates their effectiveness as tools for measuring the uniqueness of clusters.

\section{UMAP Visualisation}
\begin{figure}[H]
\centering
\includegraphics[width=\linewidth]{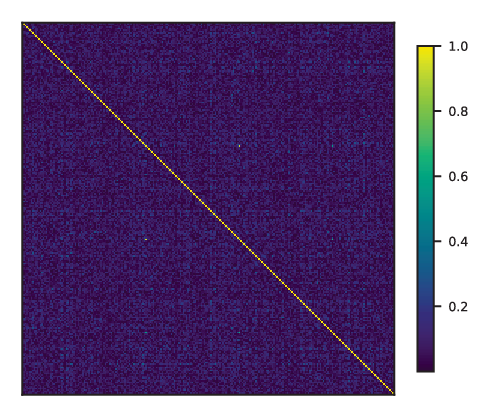}
\caption{Pearson Correlation between all 384 vectors in the LLM}
\label{fig:correlation_matrix}
\end{figure}

\begin{figure}[htbp]
 \begin{flushleft}
  \includegraphics[width=0.8\linewidth]{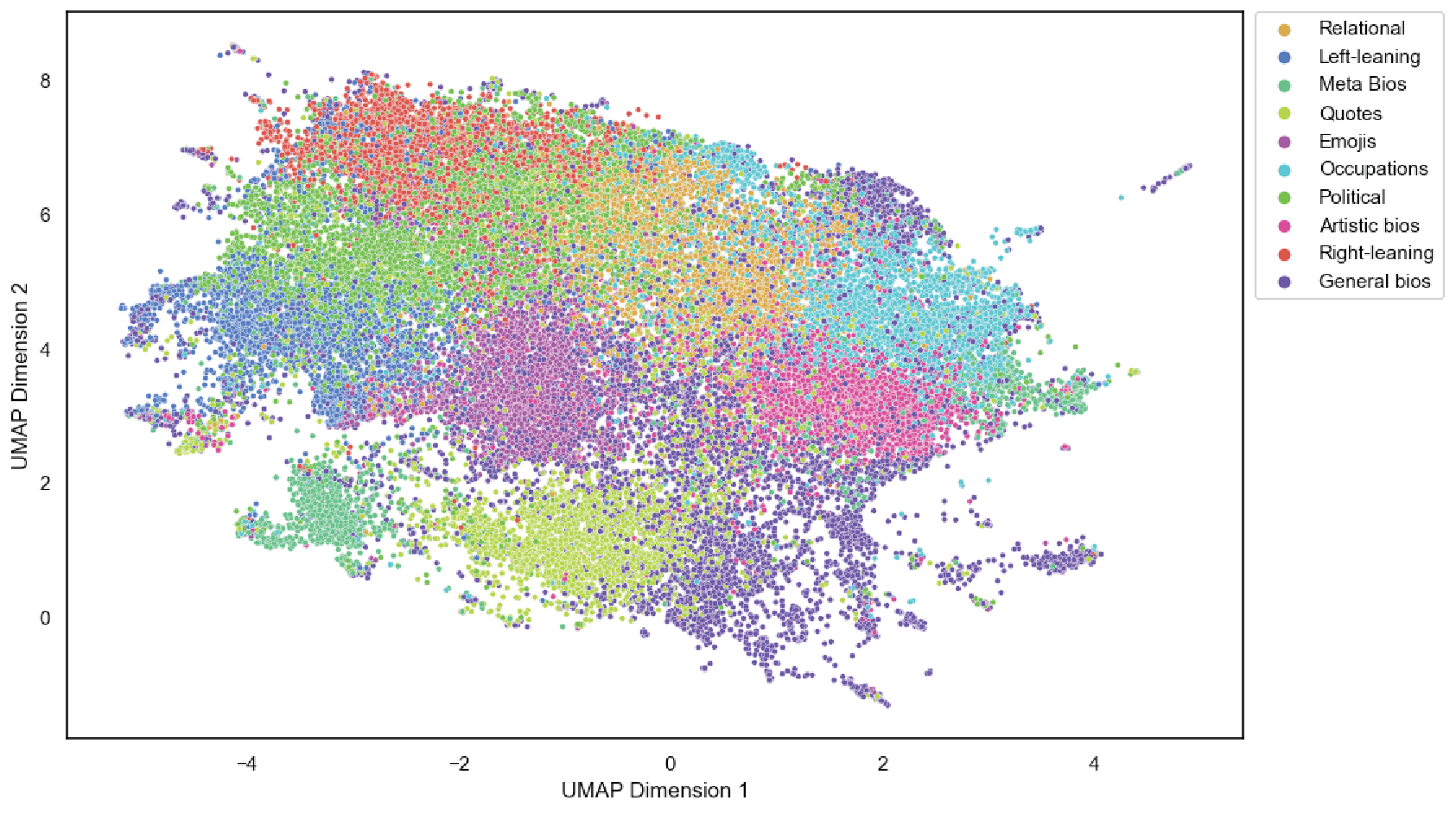}
  \caption{This figure shows all 38k Twitter bios projected onto a 2d plane using UMAP to show each points relation to each other, and which clusters are more similar to each other}
  \label{fig:UMAP}
  \end{flushleft}
\end{figure}

To effectively visualize the relationships between different clusters in our 384-dimensional large language model dataset, we employed Uniform Manifold Approximation and Projection (UMAP) for dimensionality reduction \cite{mcinnes2018umap}, see Fig.S~\ref{fig:UMAP}. The choice of UMAP is particularly pertinent due to its exceptional ability to retain the topology of the data, a feature crucial for our objective of illustrating the spatial relationships between various bios. Unlike Principal Component Analysis (PCA), which primarily captures linear relationships and variance, UMAP excels in preserving both the local and global structures within the data. This is especially important given that our dataset exhibits orthogonal dimensions, as evidenced by the correlation plot in Fig.S \ref{fig:correlation_matrix}. Although PCA might typically be effective in scenarios with orthogonal dimensions, its linear approach could potentially overlook the more complex, non-linear relationships present in our data. Therefore, UMAP's capacity to handle non-linear structures makes it a more suitable choice for our analysis, ensuring a more accurate and insightful two-dimensional representation of the clusters. The results of this can be seen in Fig.S \ref{fig:UMAP}.

\section{Stability}

\begin{figure}[H]
\centering
\includegraphics[width=0.9\linewidth]{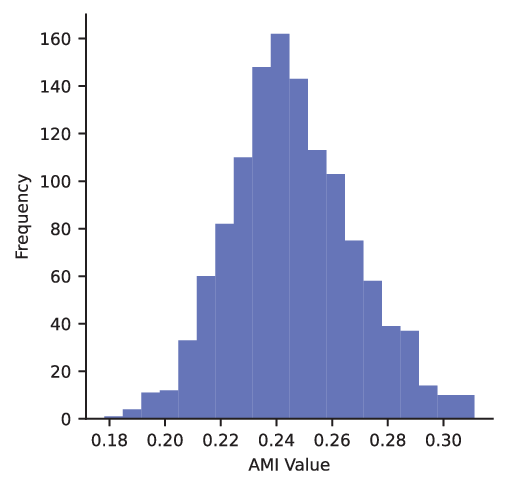}
\caption{Distribution of pairwise AMI between 50 different seeds of LDA implementation}
\label{fig:LDA_Stability}
\end{figure}

\begin{figure}[H]
\centering
\includegraphics[width=0.9\linewidth]{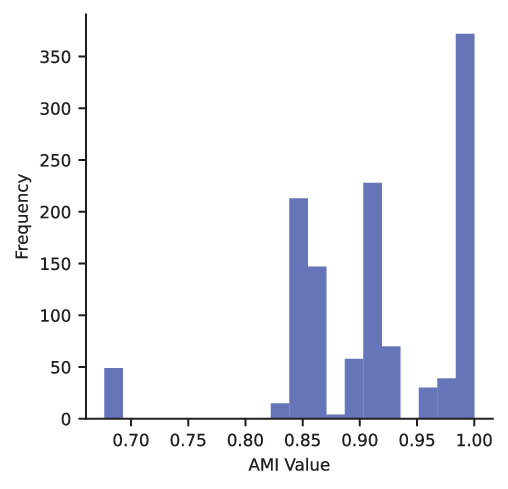}
\caption{Distribution of pairwise AMI between 50 different seeds of GMM implementation using vectors created by Do2vec}
\label{fig:Doc2vec_Stability}
\end{figure}

\begin{figure}[H]
\centering
\includegraphics[width=0.9\linewidth]{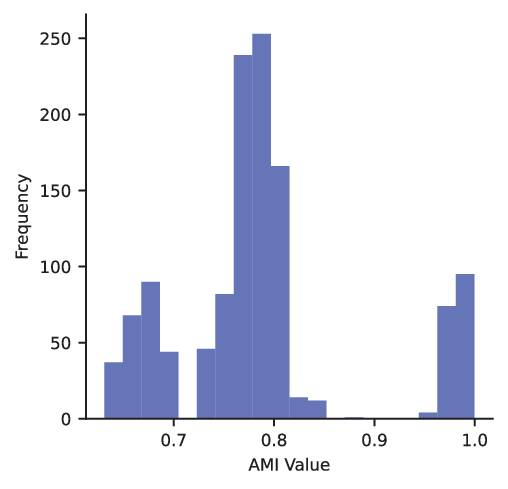}
\caption{Distribution of pairwise AMI between 50 different seeds of GMM implementation using vectors created by a LLM}
\label{fig:LLM_Stability}
\end{figure}

The three models used in the main paper each incorporate a stochastic component, which means that they can yield entirely different clusters with each run.To assess the stability of our methods, we executed them using various random seeds. For each seed, we calculated the pairwise Adjusted Mutual Information (AMI) between each seed. Our findings revealed that the LDA model had an average AMI of 0.25, with a standard deviation of 0.02. In contrast, the Doc2vec Model exhibited an average AMI of 0.91 and a standard deviation of 0.07, while the LLM showed an average AMI of 0.79 with a standard deviation of 0.09. The distribution of the AMI between each seed can be seen in  In figures S\ref{fig:LDA_Stability}, S\ref{fig:Doc2vec_Stability}, and S\ref{fig:LLM_Stability}. LDA appears to be a normal distribution. This seems to indicate that in all 50 seeds, none of them had similarly labelled clusters. This calls into question the reliability of LDA and its implementation in short text. The GMM approaches seem to be quite a lot higher, both of them seem to have around 3 peaks (a small fourth in Doc2vec), potentially indicating that there are three local maxima that the algorithm finds. All three of them seem to have a high AMI indicating they are fairly similar. However, this needs to be further explored to test what tangible differences, if any, there are between these three maxima.

\section{Other Models of ChatGPT}
\begin{figure}[htbp]
 \begin{flushleft}
  \includegraphics[width=0.9\linewidth]{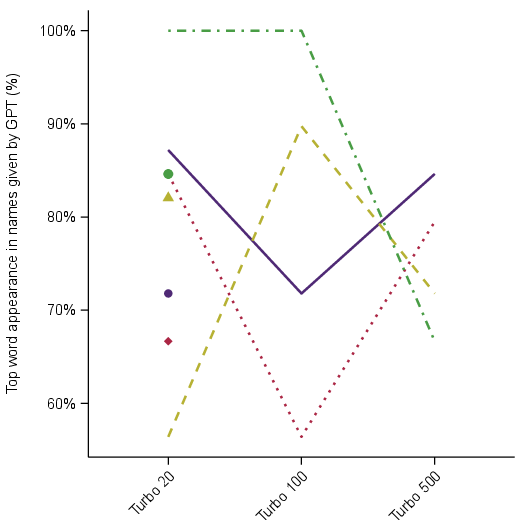}
  \caption{This figure shows the different levels of relability between gpt-4 being given 20 bios (symbols), and gpt-turbo being given 20, 100, and 500 bios, and being asked to name it.}
  \label{fig:gpt_Name}
  \end{flushleft}
\end{figure}

The use of ChatGPT allows us to explore effects which would be resource-intensive when using human reviewers.  
To evaluate ChatGPT's performance with an increased number of bios, we conducted tests using 20, 100, and 500 bios as input, focusing on clusters that represented a range in diversity. This was done to assess the model's scalability while managing costs, so we limited our tests to two clusters per method.  The procedure was consistent with our earlier experiments using the ChatGPT API, with no alterations besides the number of bios, and using GPT-4 Turbo as it can take more bios \cite{openai_2023}. After processing the bios through ChatGPT-4 Turbo, we applied the same analysis method from Section 3.8, of the method section in the main paper, to determine the frequency of the most commonly used word in the generated names. Rather than naming all clusters, we choose a small subset of clusters from across the models.  We can see in Fig.S~\ref{fig:gpt_Name} that increasing the number of bios does not appear to lead to a systematic change in the accuracy of the names produced by ChatGPT. 

\section{Process Pipeline}

\begin{figure}[H]
\centering
\includegraphics[width=0.9\linewidth]{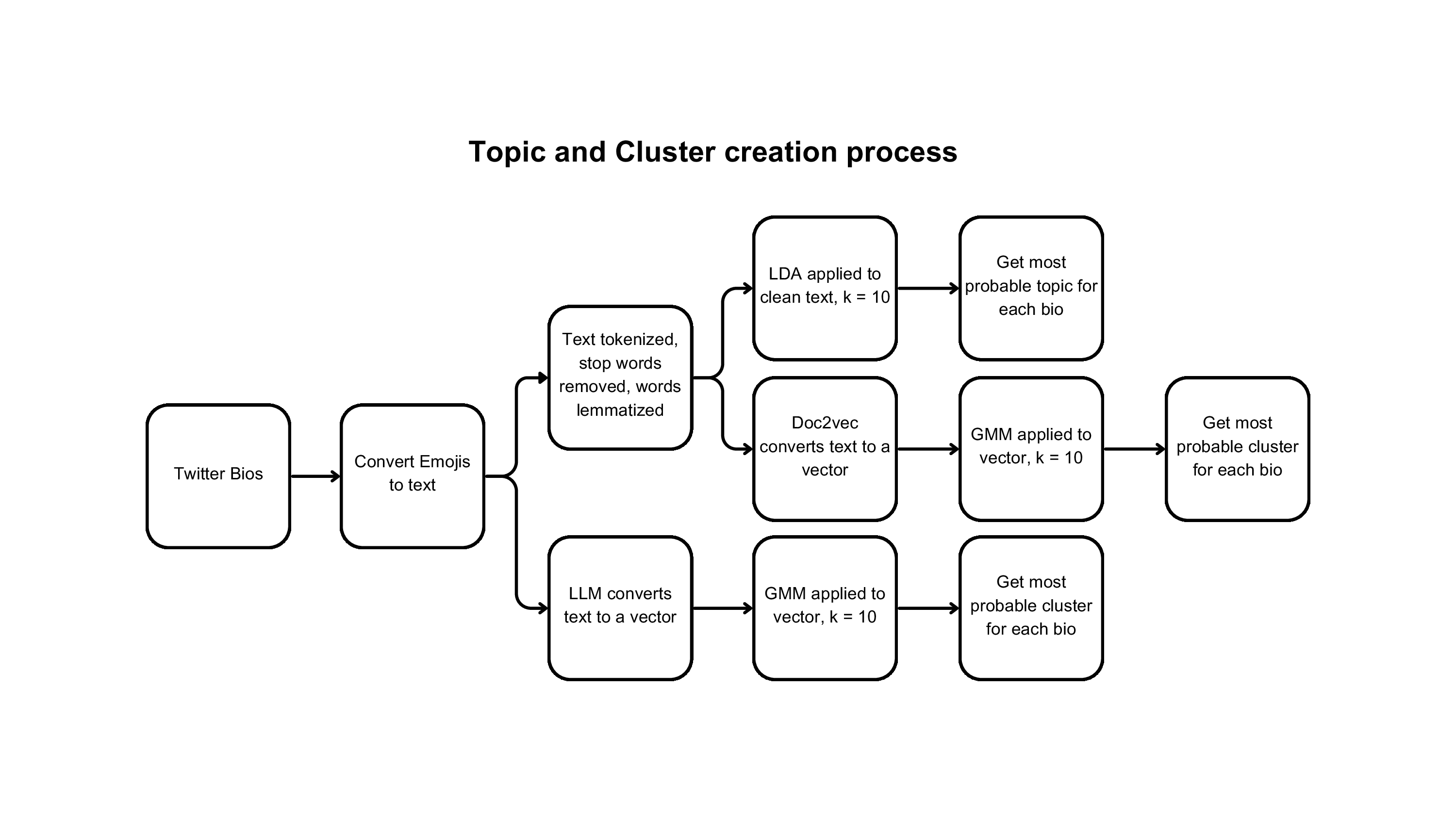}
\caption{Diagram showing the process to building each model}
\label{fig:Twitter_Bios}
\end{figure}

\begin{figure}[H]
\centering
\includegraphics[width=0.9\linewidth]{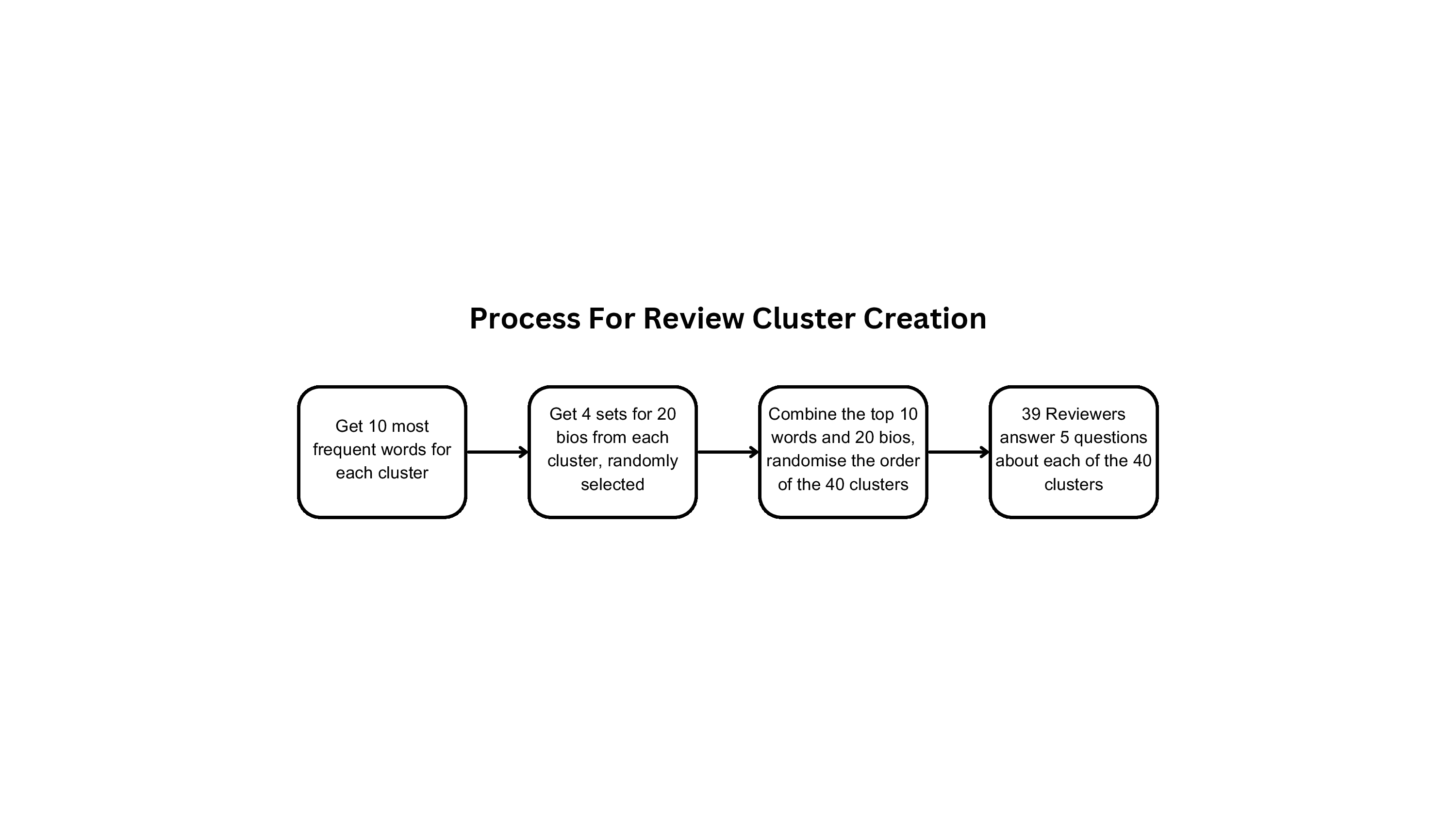}
\caption{Diagram showing the process by which the cluster samples were created to show to our reviewers}
\label{fig:Review_Process}
\end{figure}

The pipeline for how each cluster was created to show to our reviewers is in Fig.S \ref{fig:Review_Process}
The reviewers are shown each cluster in a random order and are not told which model a cluster came from. For each
of the 40 clusters, reviewers must read the top 10 words and the 20 sample bios
and then answer the following questions:
\begin{itemize}
    \item ``Create a name using less than 10 words to summarize the top 10 words/emojis and the sample bios of the Sample Cluster. If you believe it is not possible to do this with a cluster, write `None'.'' This gives a summary of whether the cluster is human-interpretable and a description of the types of bios in the cluster.
    
    \item ``When you named a Sample Cluster, were you confident that the name summarized the whole cluster?'' Reviewers can answer this question by selecting one of the following: Not at all Confident/Not Confident/Neutral/Confident/Very Confident. This question helps understand how easy it was to understand each cluster.
    
    For the following questions, reviewers can respond with: Strongly disagree/ Disagree / Neutral / Agree / Strongly Agree
    
    \item ```The situation when the top words/emojis of a cluster fit together in a natural or reasonable way' Do you agree that the above statement describes the top words/emojis of this Sample Cluster?'' This question tests whether the top 10 words/emojis of a cluster are coherent.
    
    \item ```The situation when the Sample bios of a cluster fit together in a natural or reasonable way' Do you agree that the above statement describes the sample bios of this Sample Cluster?'' This question tests whether the sample bios of each cluster are coherent.
    
    \item ```The Top Words/Emojis provide an accurate summary of the sample bios' Do you agree that the above statement describes this Sample Cluster?'' This tests if the top words/emojis and sample bios are coherent with each other.
\end{itemize}

The data given by the reviewers was then ingested and cleaned. The responses to the questions above were converted to a 1-5 scale, where a 5 represents ``Strongly Agree" or ``Very Confident" and a 1 represents ``Strongly Disagree" or ``Not at All Confident". For the confidence in naming question, some reviewers were unable to name the cluster but gave it a score of 4 or 5. In these cases, the data was changed so that every cluster that had ``None" for its name was given a score of 1 for the question on confidence in naming.

\begin{figure}
\includegraphics[width=0.9\linewidth]{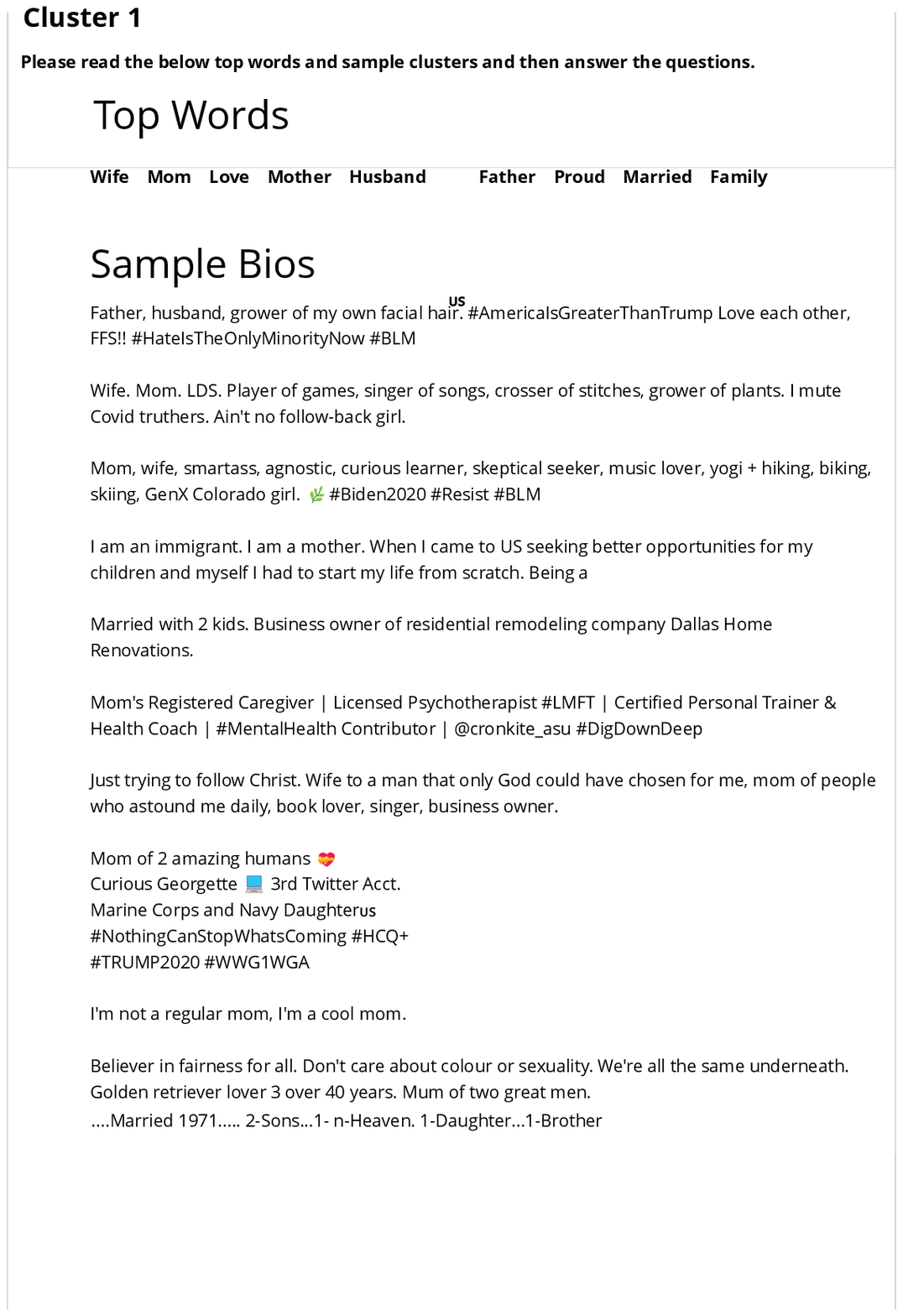}
\caption{Screenshot from Redcap of the reviewer being shown the contents of each cluster}
\label{fig:Screenshot 1}
\end{figure}

\begin{figure}
\includegraphics[width=0.9\linewidth]{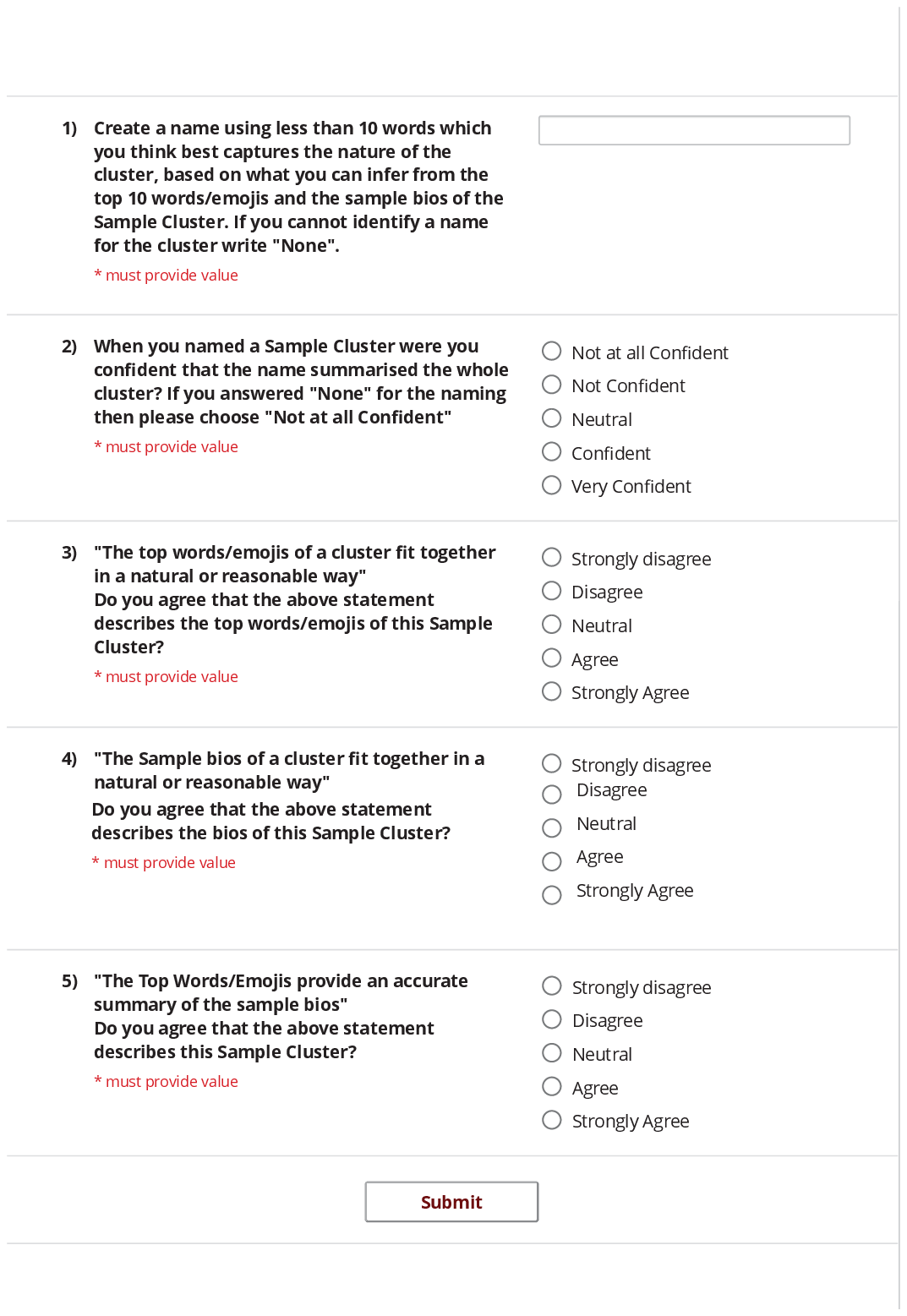}
\caption{Screenshot from Redcap of the reviewer being shown the questions and where they input their answers. Same cluster as the previous figure}
\label{fig:Screenshot 2}
\end{figure}

This study had ethics approval from the University of Sydney. Reviewers were sourced from Prolific and responses were collected through Redcap. We collected 40 reviewers, however, had to exclude one due to them misunderstanding the prompt and copying and pasting bios as the names for each cluster, as well as a non-serious attempt at the other questions. An example of how a reviewer saw each cluster can be seen in figures S\ref{fig:Screenshot 1} and S\ref{fig:Screenshot 2}

\section{Inter-Coder Reliability}
Inter-coder reliability measures the agreement between reviewers which is important given the subjective nature of their task. If there is low inter-coder agreement, this calls into question the interpretability of the clusters regardless of any other analysis. Since the data is ordinal and reviewers rate every single cluster, the appropriate method to use is the Intraclass Correlation Coefficient (ICC) \cite{gisev2013interrater, mcgraw1996forming}. The ICC is a number between 0 and 1, with a higher number indicating greater agreement between reviewers. To provide context for the ICC scores, Koo et al. provided a guideline \cite{koo2016guideline}: a score less than 0.5 is poor, a score between 0.5-0.75 is moderate, a score between 0.75-0.9 is good, and a score above 0.9 is excellent.

ICC estimates and their 95\% confident intervals were calculated using R version 4.2.2 \cite{rsoftware2022} and the 
statistical package Psych version 2.2.9 \cite{psych2022} based on an average rating, two-way random effects, absolute agreement, multiple raters/measurements. Table S\ref{table:1} shows the results of the results for each of the questions reviewers were asked.

\newgeometry{left=0.5cm, right=0.5cm}
\begin{table*}
\small
  \centering
  \begin{tabular}{|l|l|l|l|l|l|l|l|}
    \hline
    \multirow{2}{*}{Metric} &
      \multicolumn{1}{c|}{Intraclass Correlation} &
      \multicolumn{2}{c|}{95\% Confidence Interval} &
      \multicolumn{4}{c|}{F Test with True Value 0} \\
      \cline{3-8}
      & & Lower Bound & Upper Bound & Value & df1 & df2 & Sig \\
    \hline
    Confidence in Naming & .900 & .0849 & .94 & 13.5 & 39 & 1482 & \num{9.3e-73} \\
    \hline
    Coherence of Top Words & .922 & .883 & .953 & 14.7 & 39 & 1482 & \num{1.3e-79} \\
    \hline
    Coherence of Sample Bios & .911 & .866 & .946 & 13.7 & 39 & 1482 & \num{1.5e-73} \\
    \hline
    Coherence between Top Words and Sample Bios & .926 & .889 & .955 & 15.7 & 39 & 1482 & \num{7.1e-85} \\
    \hline
  \end{tabular}
  \caption{ICC Calculation in R Using Two-way random, average measure (ICC(2,k))}
  \label{table:1}
\end{table*}
\restoregeometry
From this it can be seen that the ICC is considered excellent across all four questions, indicating that the reviewers were in relative agreement on their ratings of each cluster.  A limitation of this method is that it assumes that the data is continous, but the reviewer responses are ordinal.

\section{Automated Methods}

\subsection{Automated Metric Results}
The results for every cluster's automated metrics can be found in table S\ref{table:Automated_table}
Using the automated measures to judge and compare each of the 4 models, every automated metric except for  $C_{UMASS}$ agrees with the ranking given by human reviewers: LLM, LDA, Doc2vec, and Random. However, when looking at the ranking of the individual clusters there is a much greater difference. Overall it seems that all the automated methods fail when looking at individual clusters, but in a small sample of 4 models can rank them effectively.

\subsubsection{Coherence}
Both coherence methods found that LDA 1 was the most coherent, this was in contrast to the reviewers who were not able to name the cluster and gave on average quite low scores across the 4 questions. Doc2vec had comparatively good $C_{UMASS}$ scores. In terms of $C_V$, it scored generally lower, with its highest score being for Doc2vec 4 which was one of the few that reviewers were able to consistently name. The LLM clusters that were political in nature (0,2, and 9) scored quite well on the coherence metrics, however, other clusters that were rated quite high by reviewers were given low scores. On the whole, it does not appear that either metric of coherence matches up with human interpretability.

\subsubsection{Cluster Distance}
\begin{figure*}
\includegraphics[width=0.9\linewidth]{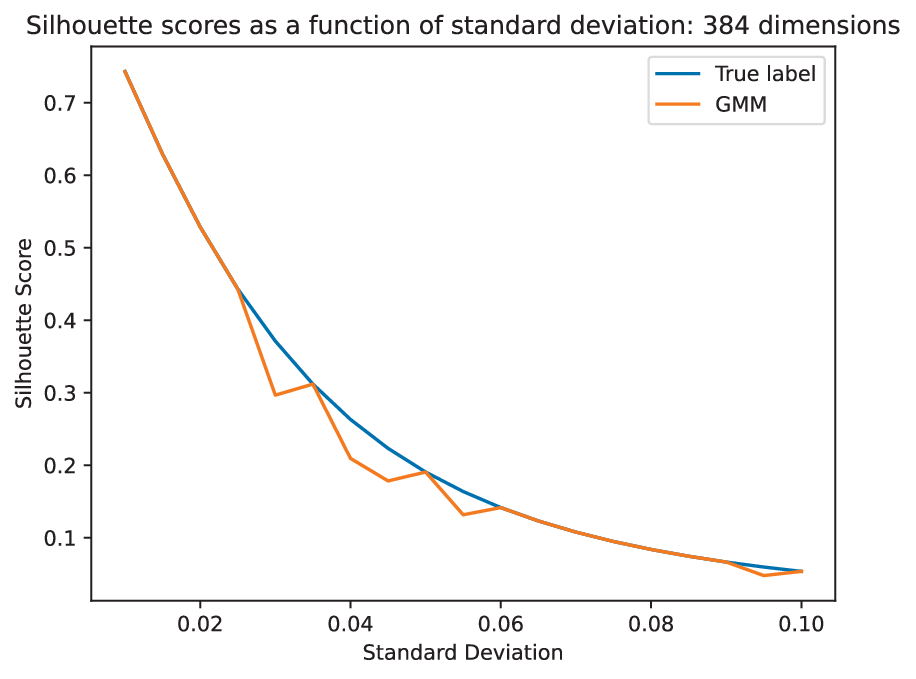}
\caption{Silhouette Score as Standard deviation increases on Synthetic data}
\label{fig:Silhouette}
\end{figure*}

The results from all models show low silhouette scores, with the majority of the scores close to zero. The reason for this is because of the high standard deviation in each dimension from the embeddings generated for the LLM and used to generate the silhouette score for all four models. Fig.S \ref{fig:Silhouette} shows that with synthetic data, the clusters generated by a GMM are able to match the silhouette score of the true clusters. As standard deviation increases it can be seen that the silhouette score decreases even amongst the true label. So the low silhouette scores are not a reflection of bad clustering but rather high standard deviation in the data. 

\subsubsection{Keyword Analysis}
Keyword analysis showed vast differences between the three models in terms of lexical content. On average, the transformer model created the clusters with the most keywords and highest average keyness scores. In more practical terms, this implies that the transformer model created the clusters that were most distinct from the original corpus of bios in terms of word frequency. In contrast, Doc2Vec was by far the worst-performing model, including several clusters with no keywords. This implies that the distribution of words in said clusters is near identical to the original corpus. Therefore, to make clusters that are lexically distinct from one another, the Transformer model performs the best.

\newgeometry{left=0.5cm, right=0.5cm}
\begin{table}[!ht]
\small
    \centering
    \begin{tabular}{|p{2cm}|p{2cm}|p{6cm}|p{1.2cm}|p{1cm}|p{1cm}|p{1.1cm}|p{1.3cm}|p{1cm}|}
    \hline
        Cluster id & Name & GPT Name & Keywords & CV & UMASS  & Distance & Silhouette & Mean SD \\ \hline
        Doc2vec 0 & U1 & Political Views and Personal Attitudes & 0 & 0.26 & -3.5 & 0.012 & -0.033 & 0.048 \\ \hline
        Doc2vec 1 & U2 & Minimal Information MAGA Supporters & 32 & 0.2 & -2.1 & 0.053 & 0.004 & 0.047 \\ \hline
        Doc2vec 2 & U3 & Diverse Personal Interests and Beliefs & 5 & 0.29 & -4 & 0.012 & -0.035 & 0.047 \\ \hline
        Doc2vec 3 & U4 & Varied Interests and Political Opinions & 62 & 0.33 & -2.6 & 0.022 & -0.024 & 0.047 \\ \hline
        Doc2vec 4 & Political\_1 & Mixed Political Sentiments and Personal Interests & 8 & 0.45 & -3.3 & 0.002 & -0.021 & 0.046 \\ \hline
        Doc2vec 5 & U5 & Mixed Political Views & 0 & 0.26 & -3.5 & 0.005 & -0.026 & 0.047 \\ \hline
        Doc2vec 6 & Political\_2 & Mixed Political Sentiments & 0 & 0.34 & -3.6 & 0.003 & -0.023 & 0.046 \\ \hline
        Doc2vec 7 & U6 & Diverse Political Interests & 0 & 0.27 & -8.3 & 0.004 & -0.010 & 0.046 \\ \hline
        Doc2vec 8 & U7 & Personal Interests and Political Affiliations & 0 & 0.44 & -8.8 & 0.004 & 0.011 & 0.045 \\ \hline
        Doc2vec 9 & U8 & Political Perspectives and Personal Interests & 0 & 0.42 & -4.4 & 0.002 & -0.022 & 0.046 \\ \hline
        LDA 0 & Left leaning\_1 & Activists and Biden Supporters & 38 & 0.35 & -2.7 & 0.086 & -0.015 & 0.045 \\ \hline
        LDA 1 & U9 & Personal Perspectives and Interests & 42 & 0.69 & -4 & 0.082 & -0.050 & 0.047 \\ \hline
        LDA 2 & U10 & Diverse Interests and Professions & 39 & 0.25 & -2.9 & 0.037 & -0.060 & 0.048 \\ \hline
        LDA 3 & Right Leaning & Patriotic Trump Supporters & 39 & 0.62 & -2.7 & 0.115 & 0.107 & 0.042 \\ \hline
        LDA 4 & Relationships & Family and Occupation Focus & 58 & 0.36 & -2.8 & 0.076 & -0.005 & 0.045 \\ \hline
        LDA 5 & U11 & Opinion Expressers & 54 & 0.42 & -3.3 & 0.049 & -0.047 & 0.048 \\ \hline
        LDA 6 & U12 & Veterans and Varied Interests & 32 & 0.31 & -2.3 & 0.043 & -0.059 & 0.047 \\ \hline
        LDA 7 & U13 & Varied Interests and Political Engagement & 51 & 0.31 & -2.7 & 0.052 & -0.039 & 0.046 \\ \hline
        LDA 8 & U14 & American Sociopolitical Views and Beliefs & 55 & 0.35 & -2.8 & 0.061 & -0.056 & 0.046 \\ \hline
        LDA 9 & Left leaning\_2 & Life, Politics, and Personal Interests & 60 & 0.35 & -2.6 & 0.037 & -0.062 & 0.047 \\ \hline
        LLM 0 & Left Leaning & Democratic Supporters and Resisters & 131 & 0.54 & -2.6 & 0.222 & 0.028 & 0.041 \\ \hline
        LLM 1 & Artistic Bios & Creative Professionals and Enthusiasts & 230 & 0.4 & -3.2 & 0.246 & 0.033 & 0.044 \\ \hline
        LLM 2 & Political & Political Affiliation and National Pride & 155 & 0.56 & -2.8 & 0.246 & -0.029 & 0.043 \\ \hline
        LLM 3 & Occupations & Professionals and Students with Various Interests & 264 & 0.29 & -4.1 & 0.183 & 0.008 & 0.045 \\ \hline
        LLM 4 & Relational & Family-oriented Patriotic Individuals & 129 & 0.38 & -2.9 & 0.304 & 0.074 & 0.042 \\ \hline
        LLM 5 & General & Eclectic Personal Expressions & 191 & 0.37 & -7.7 & 0.222 & -0.070 & 0.048 \\ \hline
        LLM 6 & Twitter & Twitter-focused Political Commentators & 133 & 0.45 & -8.1 & 0.162 & 0.035 & 0.043 \\ \hline
        LLM 7 & Quotes & Life Philosophies and Personal Beliefs & 270 & 0.43 & -7.7 & 0.334 & -0.019 & 0.046 \\ \hline
        LLM 8 & Right Leaning & Patriotic Trump Supporters & 224 & 0.7 & -1.9 & 0.291 & 0.160 & 0.038 \\ \hline
        LLM 9 & Emojis & Polarized Political Users & 216 & 0.48 & -2.6 & 0.162 & 0.099 & 0.041 \\ \hline
        Random 0 & U15 & Political Interests and Personal Identities & 1 & 0.05 & -5.1 & 0.001 & -0.002 & 0.046 \\ \hline
        Random 1 & U16 & Diverse Political and Personal Interests & 1 & 0.08 & -4.1 & 0.001 & -0.003 & 0.046 \\ \hline
        Random 2 & U17 & Polarized Political Views & 0 & 0.15 & -5 & 0.001 & -0.002 & 0.046 \\ \hline
        Random 3 & U18 & Political Standpoints and Diverse Interests & 0 & 0.09 & -4.6 & 0.001 & -0.003 & 0.046 \\ \hline
        Random 4 & U19 & Mixed MAGA Supporters and Resistors & 0 & 0.21 & -4.2 & 0.001 & -0.001 & 0.046 \\ \hline
        Random 5 & U20 & Mixed Political Views and Personal Interests & 2 & 0.13 & -4.3 & 0.001 & -0.005 & 0.047 \\ \hline
        Random 6 & U21 & Diverse Personal Interests and Politics & 1 & 0.18 & -4.2 & 0.001 & -0.002 & 0.046 \\ \hline
        Random 7 & U22 & American Politics and Personal Interests & 0 & 0.25 & -5 & 0.001 & -0.005 & 0.047 \\ \hline
        Random 8 & U23 & Mixed Political Sentiments & 0 & 0.24 & -4.2 & 0.001 & -0.003 & 0.046 \\ \hline
        Random 9 & U24 & Political Views and Personal Interests & 0 & 0.14 & -4.7 & 0.001 & -0.001 & 0.046 \\ \hline
    \end{tabular}
    \caption{Table showing every clusters score for each automated metric, including how it was named by one example use of Chatgpt}
    \label{table:Automated_table}
\end{table}
\restoregeometry

\subsection{Naming Of Clusters}

The top words that were given to each reviewer are shown in Table S\ref{table:top_words}. The LLM model appears to have a greater range of different types of bios, where the Doc2vec model only has clusters that are related to politics, and the LDA model only has 2 that are non-political, the LLM model has 5 clusters that are non-political. It also has no clusters that were not able to be named.

\begin{figure*}
\includegraphics[width=0.9\linewidth]{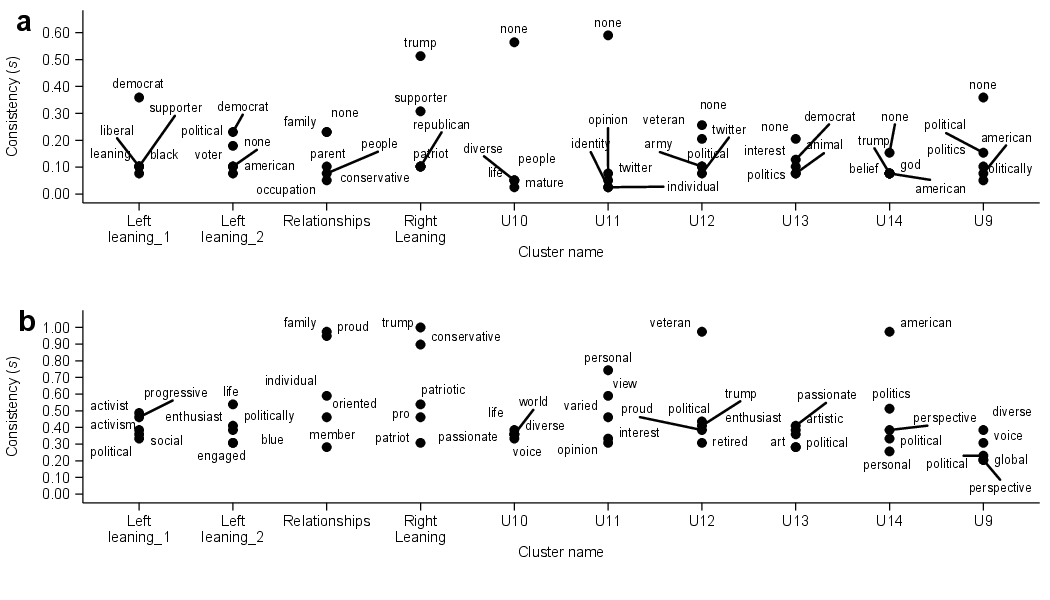}
\caption{The top five words by fraction of appearance used by (a) reviewers and (b) ChatGPT to name the clusters created by the LDA. Along the x-axis are the names given to each cluster by the authors of this paper.}
\label{fig:LDA_Names}
\end{figure*}

\begin{figure*}
\includegraphics[width=0.9\linewidth]{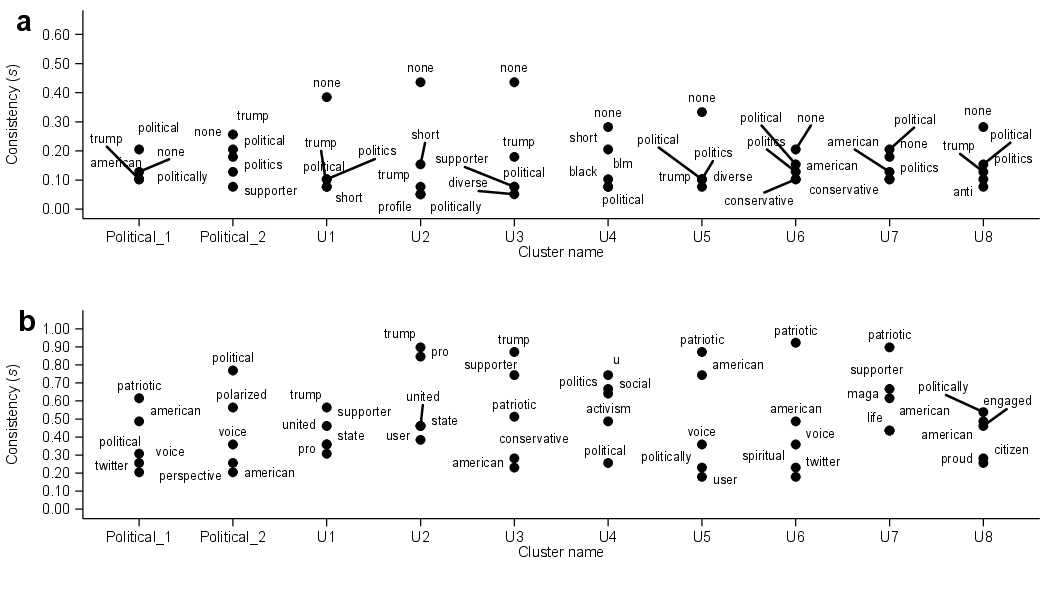}
\caption{The top five words by fraction of appearance used by (a) reviewers and (b) ChatGPT to name the clusters created by Doc2vec. Along the x-axis are the names given to each cluster by the authors of this paper.}
\label{fig:Doc2vec_Names}
\end{figure*}

\begin{figure*}
\includegraphics[width=0.9\linewidth]{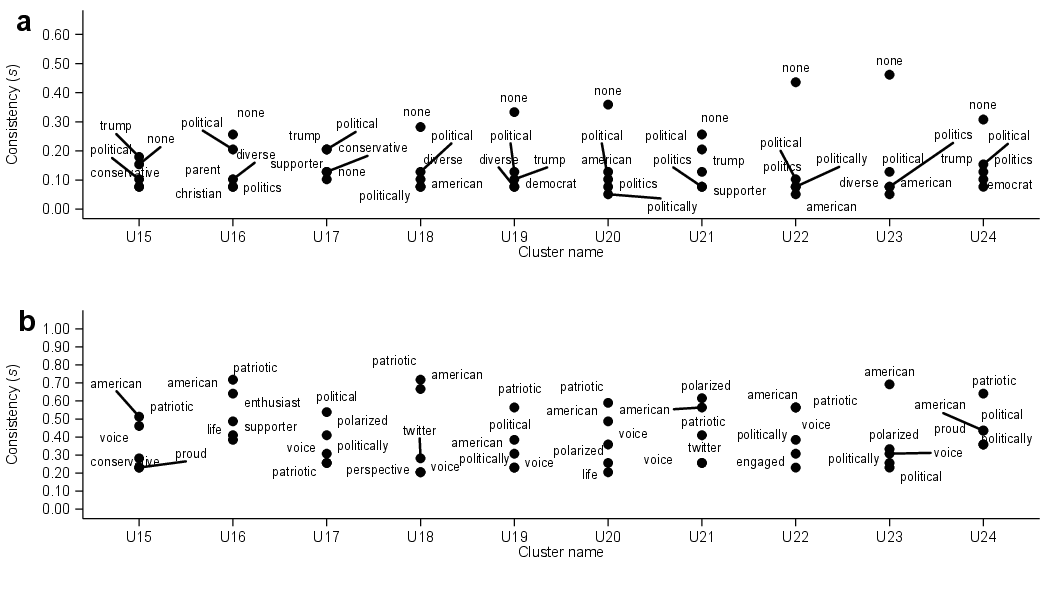}
\caption{The top five words by fraction of appearance used by (a) reviewers and (b) ChatGPT to name the clusters created by the Random Model. Along the x-axis are the names given to each cluster by the authors of this paper.}
\label{fig:Random_Names}
\end{figure*}

It can be seen that in figures S\ref{fig:LDA_Names}, S\ref{fig:Doc2vec_Names},  S\ref{fig:Random_Names}
reviewers were unable to name the LDA, Doc2vec, and Random clusters at a much lower rate than the LLM model, as evidenced by the increased use of the word "None" across those three clusters. In the clusters that were not able to be named (designated by a U), ChatGPT seems to do what the reviewers tend to do and name it based on being political. It is of note that while ChatGPT was instructed to write "None" if a cluster did not appear to have any meaning, it never did. This demonstrates a potential flaw in using ChatGPT to name clusters.

  We see that the words used to describe the clusters are largely consistent between ChatGPT and the reviewers, however there are cluster-dependent distinctions revealing human and machine limitations as discussed in the text.

From examining each model at a cluster level, the LLM model appears to create a wider range of different clusters, that are more robustly interpretable by the reviewers.

\newgeometry{left=0.5cm, right=0.5cm}
\begin{table}[h!]
\centering
\begin{tabular}{|l|l|p{13cm}|} 
 \hline
 Cluster ID & Cluster Name & Top 10 Frequent Words \\ [0.5ex] 
 \hline
 LDA 0 & Left leaning 1 & Resist, Blm, Http, Blacklivesmatter, Matter, Life, Black, Bidenharris2020, Fbr, Theresistance \\ \hline 
 LDA 1 & Undecided & New, Side, Party, Local, National, flag-mexico, Analyst, Troll, Old, General \\ \hline
 LDA 2 & U9 & sparkle, Live, Sheher, Fuck, Im, Director, Life, One, 20, World \\ \hline
 LDA 3 & Right Leaning & flag-united-states, Maga, Trump, Patriot, Kag, Love, Trump2020, God, Wwg1Wga, red-heart \\ \hline 
 LDA 4 & Relationships & Mom, Father, Husband, Wife, Retired, Proud, Mother, Lover, Fan, Dad \\ \hline
 LDA 5 & U11 & Opinion, Like, Tweet, Endorsement, Hehim, Im, La, De, Http, View \\ \hline
 LDA 6 & U12 & Army, Veteran, Vet, Retired, Follow, U, Twitter, Fan, Proud, Dont \\ \hline
 LDA 7 & U13 & water-wave, Love, Politics, News, Music, Lover, Book, Dog, World, Resist \\ \hline
 LDA 8 & U14 & Love, Trump, Truth, U, Country, Dont, God, America, Believe, Im \\ \hline
 LDA 9 & Left leaning 2 & Vote, Im, Blue, Human, Right, Life, Fan, Living, Take, Time \\ \hline
 Doc2vec 0 & U1 & flag-united-states, Maga, water-wave, Trump, Just, Trump2020, 2020, Kag, Wwg1Wga, Life \\ \hline 
 Doc2vec 1 & U2 & flag-united-states, Maga, Just, Https, Love, Wwg1Wga, Trump, Ig, Trump2020, Kag \\ \hline
 Doc2vec 2 & U3 & flag-united-states, Maga, Trump, Love, Just, Trump2020, Patriot, Https, Conservative, God \\ \hline
 Doc2vec 3 & U4 & flag-united-states, La, Blacklivesmatter, Black, Lives, Matter, Politics, Maga, Husband, News \\ \hline
 Doc2vec 4 & Political 1 & flag-united-states, Trump, Love, Maga, red-heart, water-wave, Resist, God, Life, Proud \\ \hline
 Doc2vec 5 & U5 & flag-united-states, Maga, Trump, Love, water-wave, Conservative, Proud, Life, Resist, Https \\ \hline
 Doc2vec 6 & Political 2 & flag-united-states, Trump, Love, Maga, water-wave, God, Resist, Conservative, red-heart, Kag \\ \hline
 Doc2vec 7 & U6 & flag-united-states, Trump, Maga, Love, water-wave, Proud, Resist, Mom, Life, God \\ \hline
 Doc2vec 8 & U7 & flag-united-states, Trump, Love, Maga, water-wave, red-heart, God, Proud, Life, Mom \\ \hline
 Doc2vec 9 & U8 & flag-united-states, Trump, Maga, Love, Resist, Proud, Conservative, God, Kag, Mom \\ \hline
 LLM 0 & Left Leaning & water-wave Resist, Blm, Bidenharris2020, Blacklivesmatter, Trump, Vote, Fbr, Resistance, flag-united-states \\ \hline 
 LLM 1 & Artistic Bios & Writer, Lover, Music, Artist, Love, Https, Fan, Art, Author, Life \\ \hline
 LLM 2 & Political & Trump, Conservative, Love, flag-united-states, America, Proud, American, Country, Democrat, Liberal \\ \hline
 LLM 3 & Occupations & Retired, Business, Fan, Student, Https, Science, Veteran, Politics, Engineer, Father \\ \hline
 LLM 4 & Relational & Wife, Mom, Love, Mother, Husband, flag-united-states, Father, Proud, Married, Family \\ \hline
 LLM 5 & General & Just, Https, Fan, Love, Don, Like, Life, Time, flag-united-states, La \\ \hline
 LLM 6 & Twitter & Twitter, News, Tweets, Politics, Https, Follow, Tweet, Trump, Don, Account \\ \hline
 LLM 7 & Quotes & Life, Love, God, Truth, World, People, Don, Just, Good, Matter \\ \hline
 LLM 8 & Right Leaning & flag-united-states, Maga, Trump, Kag, Trump2020, Conservative, Wwg1Wga, God, Patriot, Love \\ \hline
 LLM 9 & Emojis & flag-united-states, red-heart, water-wave, blue-heart, Love, Maga, sparkle, Trump, star, God \\ \hline
\end{tabular}
\caption{Top Words for all models}
\label{table:top_words}
\end{table}
\restoregeometry

\bibliographystyle{IEEEtran}
\bibliography{supplement}